\documentclass[journal]{IEEEtran}
\IEEEoverridecommandlockouts                              
\usepackage{graphics,balance} 
\usepackage{amssymb,hyperref,rotating}  
\usepackage{graphicx}
\usepackage{color}
\usepackage[table,xcdraw]{xcolor}
\usepackage{cite}
\usepackage{listings}
\usepackage{paralist}
\usepackage{array}
\usepackage{soul}
\usepackage{graphicx}
\usepackage{subfig}
\usepackage{floatrow} 
\usepackage{amsmath}

\newcommand{\naghme}[1]{{\color{black}{#1}}}

\title{\LARGE \bf
Masking Effects in Combined Hardness and Stiffness Rendering Using an Encountered-Type Haptic Display}

\author{Naghmeh Zamani and Heather Culbertson

\thanks{N. Zamani and H. Culbertson are with the Department of Computer Science, University of Southern California, Los Angeles, CA 90089 USA. {\tt\small naghmehz@usc.edu, hculbert@usc.edu}}%
}

\begin{document}

\maketitle

\begin{abstract}
Rendering \naghme{stable} hard surfaces is an important problem in haptics for many tasks, including training simulators for orthopedic surgery or dentistry.
Current impedance devices cannot provide enough force and stiffness to render a wall, and the high friction and inertia of admittance devices make it difficult to render free space. We propose to address these limitations by combining haptic augmented reality, untethered haptic interaction, and an encountered-type haptic display. We attach a plate with the desired hardness on the kinesthetic device's end-effector, which the user interacts with using an untethered stylus. This method allows us to directly change the hardness of the end-effector based on the rendered object. {In this paper, we evaluate how changing the hardness of the end-effector can mask the device's stiffness and affect the user's perception. The results of our human subject experiment indicate that when the end-effector is made of a hard material, it is difficult for users to perceive when the underlying stiffness being rendered by the device is changed, but this stiffness change is easy to distinguish while the end-effector is made of a soft material. These results show promise for our approach in avoiding the limitations of haptic devices when rendering hard surfaces.} 

 \end{abstract}




\section{Introduction}


\naghme{An accurate perception of hardness is essential in performing some tasks.}
For example, a dentist uses a probe to detect a soft cavity on the surface of a hard tooth; incorrect hardness perception could make the dentist misdiagnose healthy tooth as a cavity or miss a cavity completely. 
This ability to accurately detect a cavity is the main reason that dental students fail their board exam~\cite{CDCA}, highlighting the difficulty in training users for sensitive tasks in which hardness perception plays an important role. \naghme{Humans can distinguish between the hardness of different materials by tapping or pressing.}
Tactile and kinesthetic feedback~\cite{loomis1986tactual}, along with visual and audio feedback~\cite{anderson1,FLEMING201462}, can help in perceiving a material's properties.


\naghme{Haptic devices are capable of providing kinesthetic and tactile feedback, and are often used for training complex tasks such as dental procedures~\cite{gali2018technology}, orthopedic tasks~\cite{zamani21,pourkand2017mechanical}, teleoperation~\cite{van}, and device assembly~\cite{abidi2015haptics,sagardia2016platform}.
Precise hardness perception is important to accomplish these tasks.
For example, a correct perception of hardness is crucial in teleoperation in order for the operator to appropriately interact with the remote objects~\cite{van}. 
Medical haptic training simulators also require high-precision haptic force rendering to create an experience that closely matches clinical practice~\cite{aggarwal2010training,zamani2019novel}. }


The fidelity of haptic simulation is defined by how similar the virtual object is to the real object. In the real world two solid objects cannot pass through each other, but in simulation, virtual objects can interpenetrate unless some penetration constraints are required~\cite{WANG2019136}.
To render a stiff object in an ideal virtual simulation, interpenetration of rigid objects should not be allowed. This ability to render stiff virtual objects was first proposed in~\cite{Salisbury}, and has been one of the most critical challenges in haptics. 

Two significant criteria in the design of haptic displays are: (1) allowing users to move effortlessly with no friction and inertia when in free-space, where there is no virtual contact; (2) providing a high force for generating a stiff surface without losing stability. When we render a stiff virtual wall, time-discretization, quantization of encoder measurements and commanded force, and a limited bandwidth of the actuation system can create an unstable system, which might be felt by the user as an oscillation~\cite{Diolaiti}. Due to these issues we are limited in the maximum stiffness that we can stably render using a kinesthetic device.

We can render stiffness using one of the two types of haptic devices: impedance or admittance.
In impedance control devices, the input location is measured and the reaction force is fed back to the operator. Usually, these devices provide lower force but have less inertia. In admittance control devices, the forces exerted by the operator are measured and positions are fed back to the operator. These devices can provide a higher force, but have higher inertia. There is a direct trade-off between impedance and admittance devices in their ability to create free space and provide high-force feedback.

Here we list some of the current impedance haptic devices and their maximum forces given in their respective datasheets: Novint Falcon (9~N), 3D Systems Touch X (7.9~N), and Force Dimension Delta3 (20~N). The maximum stiffness that these devices can generate is not high enough to render a realistic wall. One approach to increasing force output is to build an impedance device with more powerful actuators. However, this increases the friction and inertia, and requires the user to carry the weight all the time, making this approach impractical. 
To address rendering within the limitations of kinesthetic devices, \naghme{previous studies attempted to render hardness separately using event-based haptic rendering such as rate-hardness~\cite{Kuchenbecker1, Lawrence}, which increased perceived hardness of the virtual surfaces.} Building on this prior work, we propose that hardness must be independently rendered along with the stiffness rendered by the haptic device. With this method, we can simulate both transient- and extended-response force during the haptic interaction with objects in order to achieve true realism given the limitations of the haptic device.

\begin{figure}[tb]
   \centering
   \includegraphics[width=0.8\linewidth]{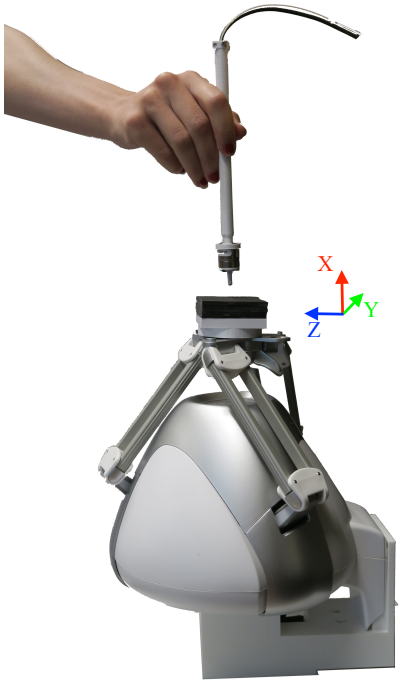}
   \vspace{-0.15in}
   \caption{Experiment Setup: kinesthetic haptic device with an unthethered stylus.}\label{fig:setup}
 \end{figure}

This paper seeks to improve the realism of rendering virtual surfaces with an impedance haptic device by combining stiffness rendered by the haptic device with hardness cues from a physical surface attached to the device's end-effector (Figure~\ref{fig:setup}). The user directly interacts with the surface on the end-effector using an untethered stylus, creating the basis of our Encountered-Type Haptic Display (ETHD). This paper discusses the design and control of a modified off-the-shelf haptic device, and studies how hardness of end-effectors can mask the stiffness provided by the device.

\section{Background}

\subsection{Hardness and Stiffness Perception}
Research into objects' perceptual dimensions has shown that hardness is a key physical property~\cite{okamoto2012psychophysical}. Often an object's hardness and stiffness are lumped together in haptic rendering, but these are two distinct properties that uniquely contribute to the perception of the object. 
An object's stiffness can be represented as its compliance, which is the ratio between an applied force and the resulting displacement. A softer object has a lower stiffness, which depends both on the object's material and physical dimensions. For example, a thicker object can be compressed more than a thinner object of the same material when the same force is applied. An object's compliance can also be represented by the material's Young's modulus, which is the ratio between pressure and displacement, and is independent of the object's dimensions~\cite{wouter1}. 

Harper and Stevens found that the perceived stiffness and hardness for natural objects are related through a power function with an exponent of 0.8~\cite{Roland}. Local surface deformation in these objects provides more critical information about the hardness than stiffness. 
A later study found that hardness perception is likely to be based on the relationship between the force acting on the object surface and its surface vibration~\cite{Higashi1}. Due to this relationship, tactile cues play an important role in hardness perception, unlike for stiffness. 
Haptic devices provide only global force and motion, and they do not provide any sensory information about local object deformation or surface vibration cues. Consequently, for rendered objects, the relationship between object stiffness and perceived hardness is not as straightforward as for natural objects~\cite{Lawrence}. 

Previously, stiffness was used to predict the perceived hardness of a virtual surface, but this measure was not accurate due to the fundamental difference between stiffness and hardness. The perceived hardness is related to the dynamics of the interaction rather than the static object stiffness. 
Rate-hardness, the initial rate of change of force versus velocity upon penetrating the surface, has been introduced as a more accurate measurement of perceived hardness than stiffness~\cite{Lawrence}. 



The majority of studies on hardness perception have focused on bare-hand interactions (for example, see~\cite{tiest2009cues,Friedman,srinivasan1995tactual}), despite the importance of tool-mediated interaction in current force-reflecting haptic interfaces~\cite{Kuchenbecker1,Han}. 
Understanding hardness perception in tool-mediated interactions is essential for designing haptic interfaces and rendering algorithms.

Peak force and peak force rate were previously shown to be essential parameters for hardness perception~\cite{Okamura1}, but later studies revealed that they are irrelevant to hardness perception through a rigid tool~\cite{Han,Friedman}. In~\cite{Han}, they found a high correlation between the peak force and peak force rate averaged across tapping velocities, but their actual correlations to perceived hardness were low. Therefore, they created a better measurement of perceived hardness by extending the rate-hardness to consider the steepest force change rather than only the initial force after a contact that occurs within a short time interval. This new measurement, the extended rate-hardness, has a significantly  higher correlation with the hardness perceived through a rigid tool. 

\subsection{Hardness and Stiffness Rendering}
Researchers have worked to increase the impedance range of haptic devices using proper kinesthetic cues in order to increase the force and realism of the haptic interaction, while maintaining stability. A virtual coupling algorithm was one of the earliest methods, in which the actuator force is restricted with respect to the penetration depth and generated energy in order to guarantee stability~\cite{Colgate1}. The computation load in this method is high and a very stiff object could not be rendered. 

{Passivity theory has been used to provide stability conditions in kinesthetic devices~\cite{colgate1997passivity,hannaford2002time}. It has been shown that the zero-order-hold creates a delay that generates energy into the system, which causes instability if it does not dissipate through friction of the device or the control system. This energy is proportional to the stiffness of the virtual object~\cite{colgate1993implementation,gillespie1996stable}.  
In~\cite{Jong}, authors considered the amount of energy dissipated by the physical damping of a device and used the passivity approach to bound the output energy. Even though the system was stable, and it was stiffer than the virtual coupling method, the perceived stiffness was lower than the desired stiffness. Singh et al. created a novel rendering algorithm that increased stiffness, but which also decreased rate-hardness~\cite{Singh}. They later improved their approach to increase stiffness while also enhancing the rate-hardness~\cite{Singh2}.

During contacts with rigid objects an initial high-frequency transient force occurs in addition to a slower extended response force. Generating force-transients using closed-loop control is not sufficient and does not achieve true realism because it does not display the faster transients that are essential for stimulating the user’s perception~\cite{Kuchenbecker1}. An event-based approach solves this issue by using kinesthetic haptic devices to cancel the incoming momentum of the user’s hand during initial contact with the virtual object.
These high-frequency contact transients have been shown to improve the realism and hardness of a virtual surface~\cite{Kuchenbecker1,Okamura1,Han2}.
Although this method improves hard object rendering, there is a limit to the strength of the transients that can be displayed as well as the stiffness of the underlying surface due to the current limit of the haptic device's motors.
To increase the realism of event-based interactions for harder objects like metal, devices with higher peak current capacity are required~\cite{Kuchenbecker1}. Also, since there is no proper physical measurement, it is difficult to understand how to simulate an object's exact hardness and how close we can come to matching the real hardness. }

In recent years, researchers have also worked to increase perceived hardness by including tactile contact cues, which have been found to influence perceived hardness more than kinesthetic cues~\cite{Tiest1,Okamura1,Kuchenbecker1,Han2}. For example, researchers have shown that adding cutaneous feedback for both one-finger touch and two-finger grip of virtual objects can increase hardness perception of the virtual objects~\cite{Park}.
Pourkand et al. also worked to increase perceived hardness by creating a haptic illusion using a hybrid force-moment braking pulse~\cite{Pourkand}. Research has shown that adding damping to the rendered surface can be effective in increasing the perceived hardness of virtual objects only if contact with the virtual surface is maintained~\cite{van}. During contact-transition tasks, added damping decreases perceived hardness and has a more substantial negative effect than its improvement to perceived hardness during in-contact interactions.

Haptic augmented reality (AR) is another research area that has been implemented to improve the realism of virtual objects. For example, \naghme{in~\cite{Jeon} the authors created }a haptic AR framework, in which the stiffness of a real 3D object can be modulated with additional virtual force feedback. Limitations of this method include that it can only operate with real objects that have moderate stiffness and it assumes the real objects have a homogeneous dynamic responses. Highly stiff objects cannot be used due to the limited performance of current haptic devices. Although this method can be beneficial, it cannot fully solve the virtual wall problem due to its limitations.

\subsection{Encountered-Type Haptic Displays}
The most common method for interacting with kinesthetic haptic devices is through a stylus or similar object that is connected to the end-effector of the device. The user directly moves the end-effector, which senses their motion and controls the user's position in the virtual environment. This direct connection means the user has to be in contact with the device's actuators, which adds unwanted friction and inertia to the interaction, including when simulating free space.

In ETHDs the user only touches the end-effector when in contact with a virtual object, which can be a solution for this problem because the user is naturally untethered in free space.
A review of previous grounded ETHDs is presented in~\cite{Rodrigo}. For example, the device in~\cite{snake} attaches different textures to the surfaces on the end-effector of a robot. The appropriate texture is presented to the user when they contact a virtual object to create the illusion that the object is physically there.
Also in~\cite{Mercado} they created an ETHD using a cylindrical spinning prop that is attached to a robot’s
end-effector. Different textures, including a wooden desktop, a leather cover, and a sheet of paper, were attached to the cylinder so they could switch between them in VR.

These previous ETHDs have been created using admittance robotic arms, and their purpose was to create on-demand tangibles in VR. We seek to use an ETHD to increase the realism of virtual haptic interactions by matching the hardness of real objects.
This paper presents our system for simultaneously rendering hardness and stiffness using an off-the shelf impedance haptic device.





\section{Encountered-Type Haptic Display Design}
Building on prior work in ETHDs, we propose a novel system for rendering both stiffness and hardness. 
Due to the decoupled nature of ETHDs, we have a virtual stiffness (controlled by the haptic device) and a real hardness (due to the collision impact between the stylus and the device's end-effector).  Here we present our design, in which we transform a traditional kinesthetic haptic device to act as the base for an ETHD system.


\subsection{Hardware Design}

We modify a Novint Falcon haptic device by detaching the stylus from the body of the device and use the end-effector as the primary interaction point between the stylus and the body, as shown in Figure~\ref{fig:setup}. Surfaces of different hardnesses can be directly attached to the end-effector, as discussed below. {The Falcon device is an inexpensive haptic device that has limited force rendering capacity, making it impossible to perfectly render a stiff material.} We chose to work with the Falcon device to test our methods on improving the rendering capabilities of a device with limited capabilities; our methods can easily be applied to other kinesthetic devices.
The maximum force that this device can provide is in its X-direction, normal to the body of the device. Since our goal is to study the relationship between stiffness and hardness, we expect the primary interaction behavior of our users will be tapping on the surface~\cite{lamotte2000softness}. Therefore, we rotate the device 90 degrees to maximize the output force in the vertical direction. We then add a new gravity compensator to our system in the X-axis to negate the effect of the weight of the end-effector.

The first benefit of this design is that we can eliminate the effect of friction and inertia of the joints and links of the device while the user interacts with free space.
Secondly, the system design is not device-dependent and does not require the use of an impedance device. The kinesthetic device itself could be replaced with any device that allows vertical application of force, including an admittance device that may be able to provide higher force and stiffness.

\begin{figure}[tb]
  \centering
  \includegraphics[width=0.85\linewidth]{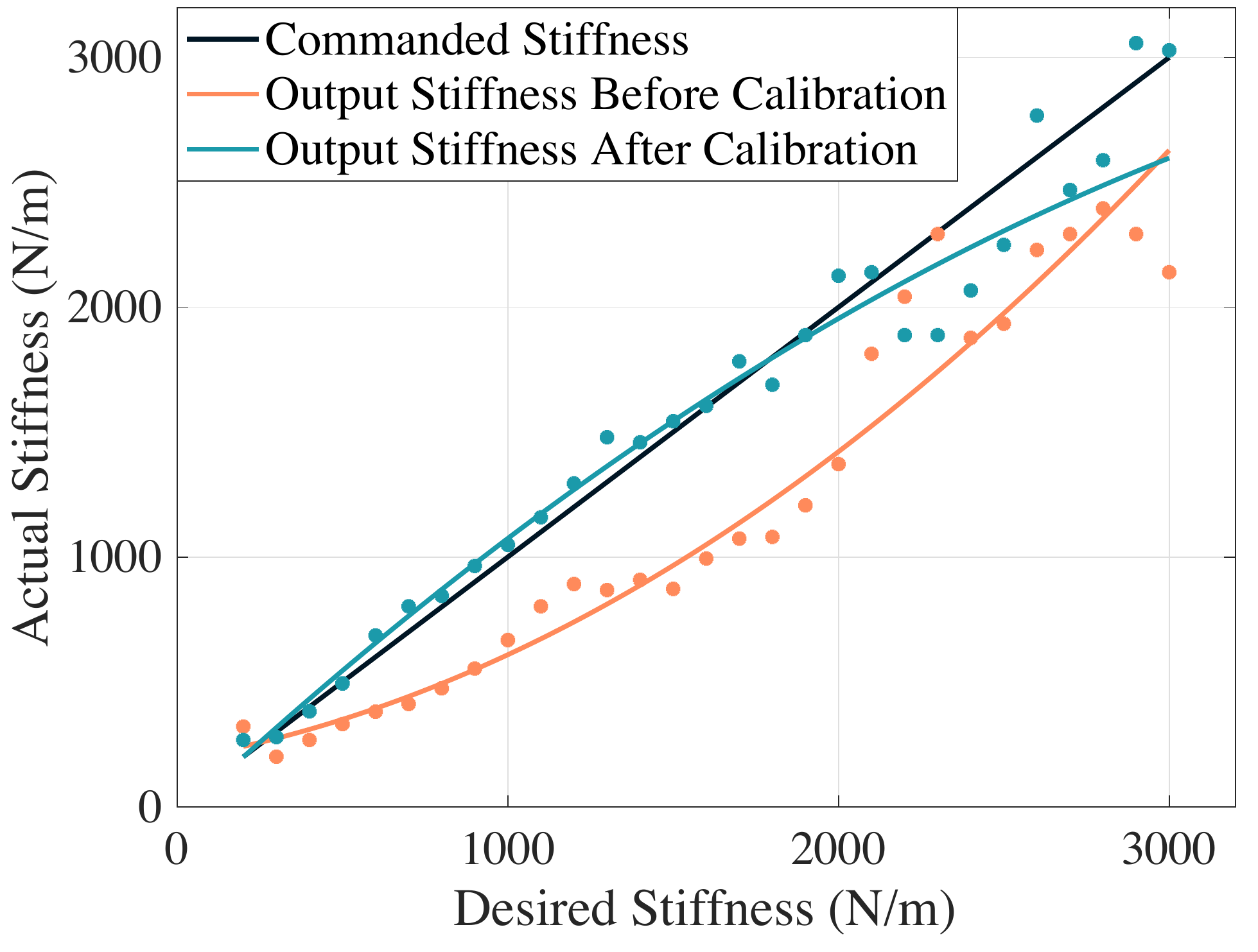}
  \caption{Desired stiffness versus actual stiffness provided by the device before and after calibration. 
  }\label{fig:calib}
\end{figure}

\subsection{Evaluation and Calibration of the Haptic Device}
Since the Falcon haptic device works in open-loop control, we need to make sure that the device's output force is close to our commanded force. We evaluated and manually re-calibrated the device after the {auto calibration that is created by the Force Dimension Company for this device.}

For this calibration we placed a constant weight (100~g) on the end-effector, and increased the stiffness from 100~N/m to 4000~N/m in increments of 100~N/m. We recorded the distance traveled by the end-effector after each increment using the device's encoders. For each stiffness, we ran this experiment and recorded the position five times, averaging the distance traveled. {Then the actual stiffness was calculated by dividing the constant weight by the average traveled distance.}

Figure~\ref{fig:calib} shows the results from the calibration. Regardless of the commanded desired stiffness, the maximum stiffness that the device was able to provide was around 2000~N/m; the plot shows a saturation after this point. Therefore, we use 2000~N/m as the maximum stiffness for our experiment, to make sure the device output is accurate.

Figure~\ref{fig:calib} shows a falsely low output stiffness for all commanded stiffness values. We fit a second-degree polynomial to this data to determine this offset. We then removed the offset using the inverse of this polynomial as the relationship between the commanded stiffness and desired output:
{\begin{equation}
k_{out} = -3.49e-04*k_{des}^2 + 2.03*k_{des} - 147.27
\end{equation}
where $k_\textrm{out}$ is the output stiffness commanded to the device, and $k_\textrm{des}$ is the desired stiffness.}

To test the stiffness calibration of the device, we re-ran the above data collection with the new commanded stiffness values. Figure~\ref{fig:calib} shows a significant improvement in the match between desired and commanded stiffness, especially in the range of interest $k<2000$~N/m.


\begin{figure}[tb]
  \centering
  \includegraphics[clip, trim=0cm 0cm 0cm 0cm,width=\linewidth]{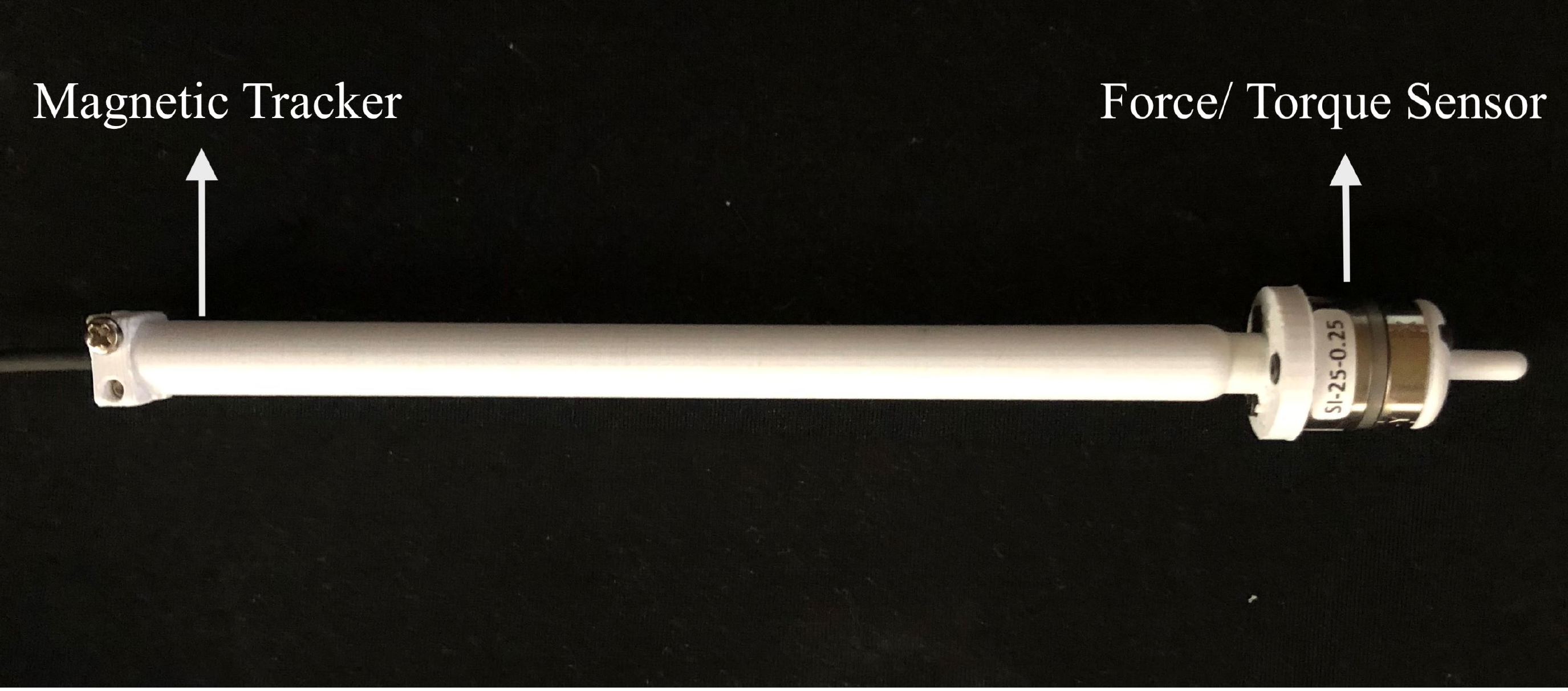}
  \caption{A 3d printed stylus with embedded position sensor and a Force/Torque sensor}\label{fig:tool}
\end{figure}

\subsection{Stylus}
In this design, the stylus is ungrounded and is separated from the body of the haptic device. \naghme{We embed sensors in the stylus so that it can detect its own location and force that applies to the end-effector. Currently we use this data to find the relationship between human perception of hardness versus speed and force of tapping, but in future this data will be used in the haptic loop for real-time adjustable rendering based on speed and force.

For this stylus, we designed and 3D printed a} 12~cm length pen-like stylus with 1~cm diameter (Figure~\ref{fig:tool}).  We embedded a Nano17 F/T sensor (ATI Industrial Automation) in the stylus to record the force of the interaction between the stylus and the end-effector. \naghme{The sensors' data was logged by a Sensoray 826 PCI Express board}. We also included an Ascension trakSTAR magnetic tracker (0.5~mm resolution) to track the location, orientation, and speed of the stylus. The magnetic tracker is embedded in the back of the stylus to avoid any interference or noise with the force sensor.  The stylus has a {4~mm spherical tip that is 15~mm in length} to ensure there is only one interaction point with the end-effector surface.

\subsection{End-effector Plate}
In this ETHD, we use the device's actuators to render stiffness and the end-effector surface to render hardness. The hardness of the end-effector can be directly altered by attaching different materials to its surface. The hardness and thickness of the materials will affect the interaction impact of the stylus with this surface. {In this work, we explore the effect of interacting with plates of different hardnesses on the end-effector, and study their influence on human perception of stiffness and the overall interaction.}

\section{Experimental Methods}
We conducted two experiments to understand the interplay between the rendered stiffness and physical hardness of the end-effector on perceived hardness of the virtual wall. We first quantitatively evaluated this effect by measuring the spectral centroid of force during tapping for varying stiffness and plate hardness. We then qualitatively evaluated the effect of stiffness and physical hardness through a human subject study.


\subsection{Experiment 1}
Previous studies agree that a significant cue in hardness perception is the frequency of the transient vibration produced during tapping~\cite{Kuchenbecker1,Okamura1,Ikeda,Higashi2}. Therefore, in this paper, we study hardness perception in our device during tapping.
For the initial experiment, we determine the effect of varying stiffness on a quantitative measure of perceived hardness, the spectral centroid of the tapping transient, which has been shown to be an effective measure of perceived hardness~\cite{Culbertson17}. In this experiment, we used a set of 11 plates with different hardness attached to the device's end-effector using a custom 3d printed attachment (Figure~\ref{fig:tip}). These plates, shown in Table~\ref{table:hardness} were 2~in$\times$2~in wide and ranged from extra soft to extra hard (with thickness of 0.5~in for soft materials, and 0.125~in for hard materials). {The hardness of these materials are specified on the Shore durometer scale, which measures the penetration depth under a specific load.}

\begin{center}
\begin{table}
\caption{{Selected hardnesses covering a wide range}}\label{table:hardness}
\begin{tabular}{ |c|c| } 
\hline
Shore No. & Range \\ 
\hline
\hline
Shore 10OO & Extra Soft \\ 
\hline
Shore 40OO & Extra Soft \\ 
\hline
Shore 50OO & Soft \\ 
\hline
Shore 60OO & Soft \\ 
\hline
Shore 70OO & Medium\\ 
\hline
Shore 40A & Medium\\ 
\hline
Shore 60A & Medium \\ 
\hline
Shore 80A & Hard \\ 
\hline
Shore 90A & Hard \\ 
\hline
Shore 95A & Extra Hard \\ 
\hline
Shore 75D & Extra Hard \\ 
\hline
\end{tabular}
\end{table}
\end{center}

\begin{figure}[tb]
        \subfloat[\naghme{Tapping force, horizontal line shows average of peaks.}]{\includegraphics[clip, trim=0cm 4.5cm 1.5cm 5.5cm,width=0.45\linewidth]{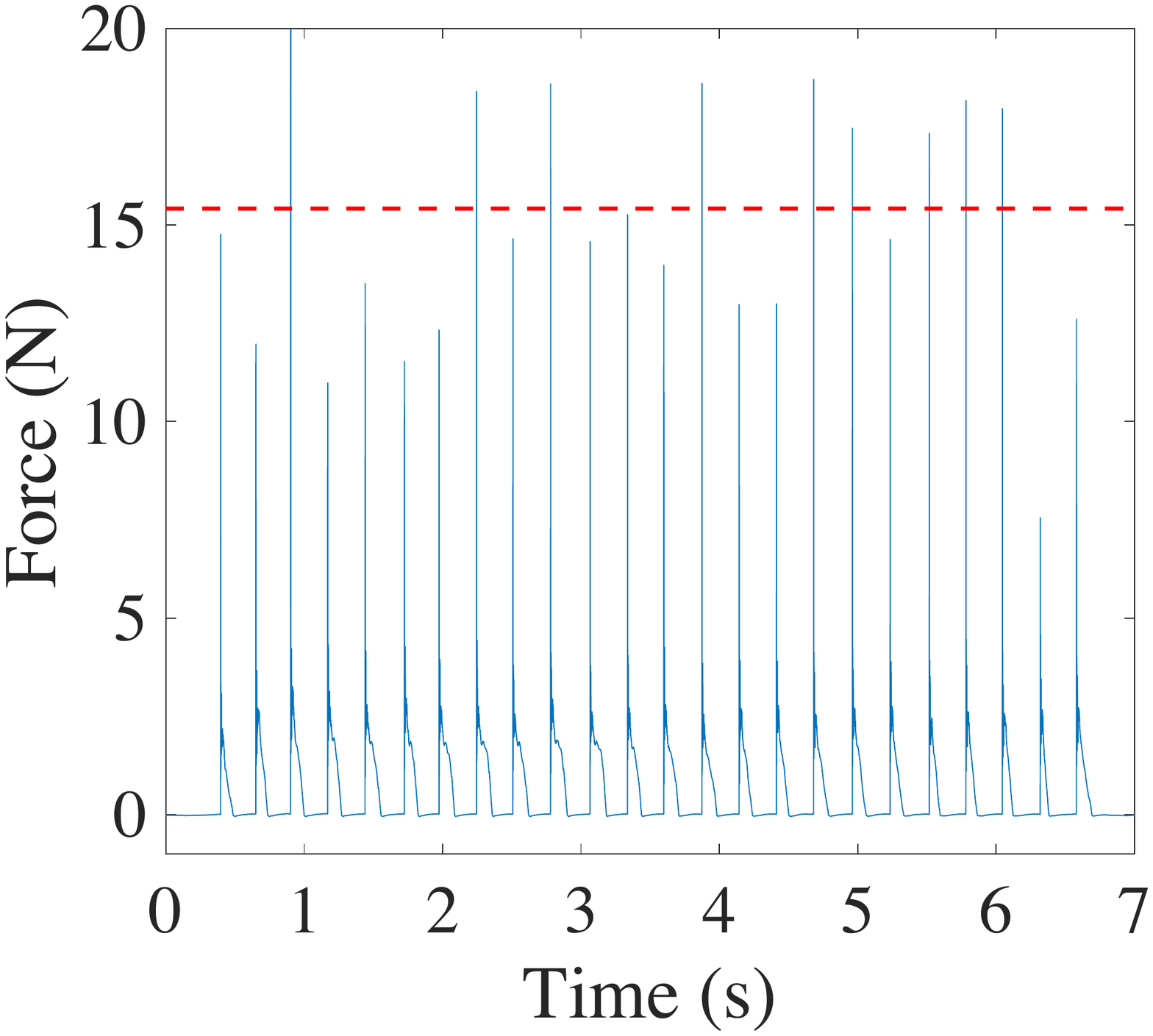}
        {\includegraphics[clip, trim=0.cm 4.5cm 1.5cm 5.5cm, width=0.45\linewidth]{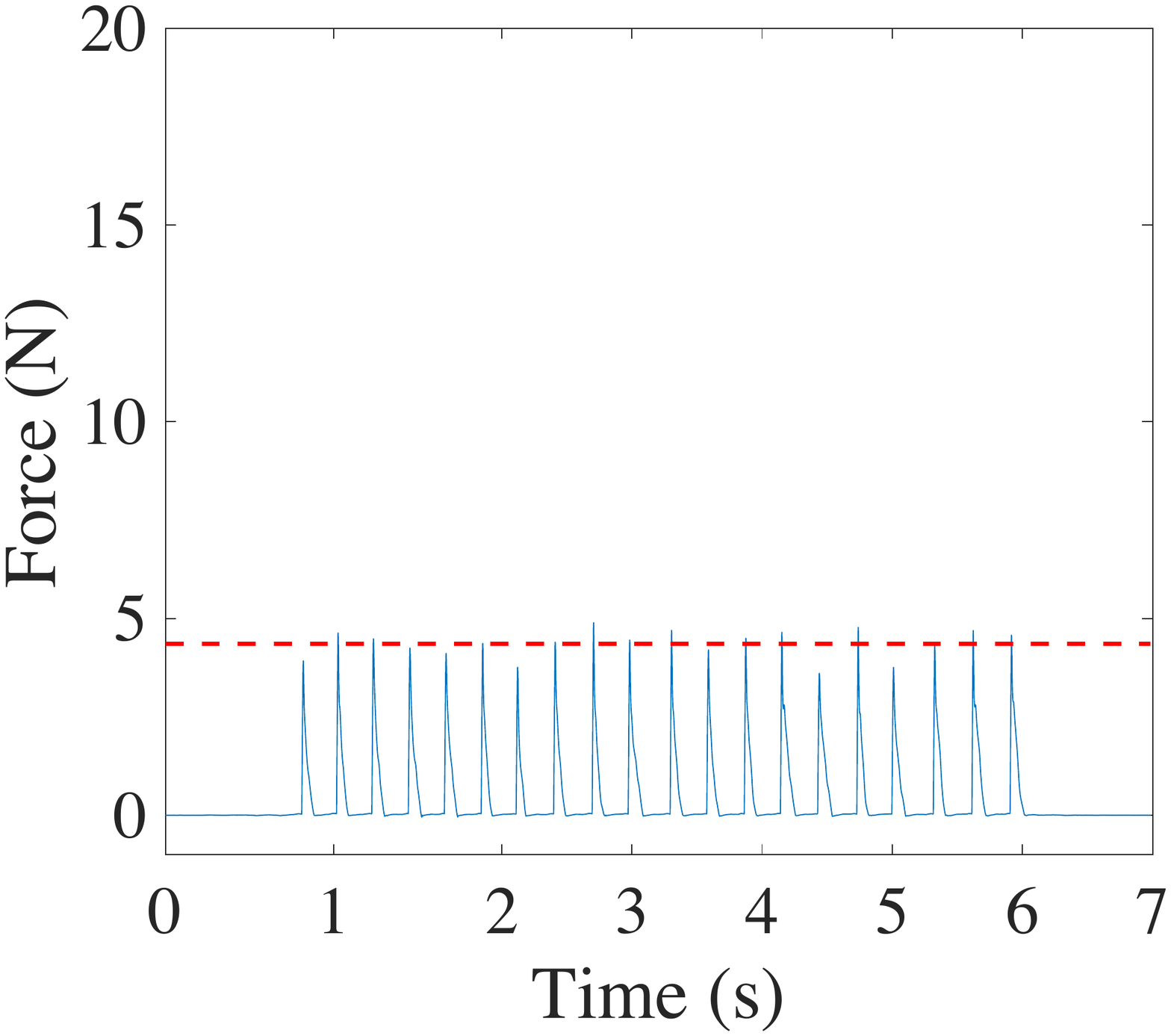}}}\\
        \subfloat[\naghme{Selected closest tap to the vertical average line.}]{\includegraphics[clip, trim=0.cm 4.5cm 1.5cm 5.5cm,width=0.45\linewidth]{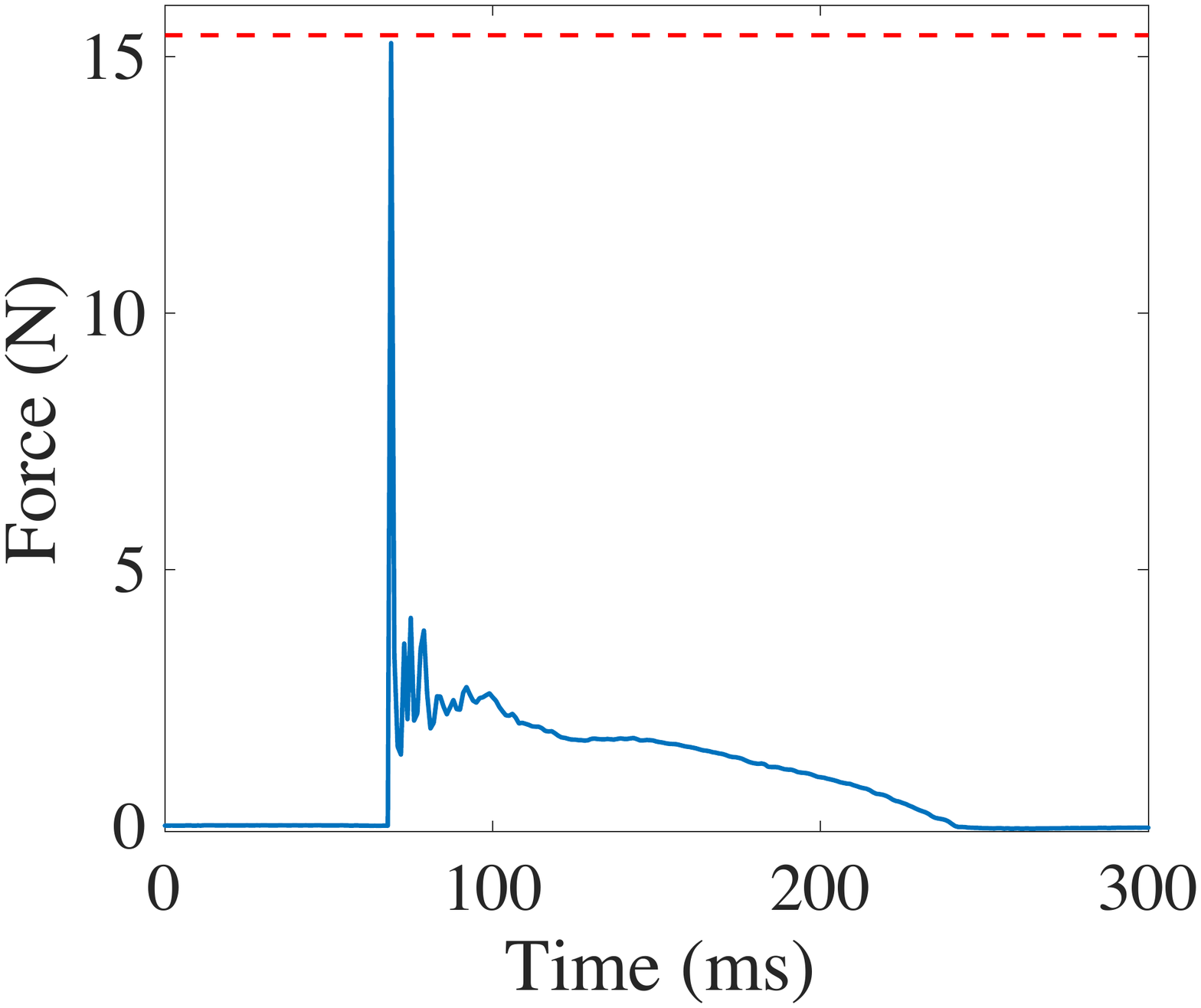}
        {\includegraphics[clip, trim=0.cm 4.5cm 1.5cm 5.5cm, width=0.45\linewidth]{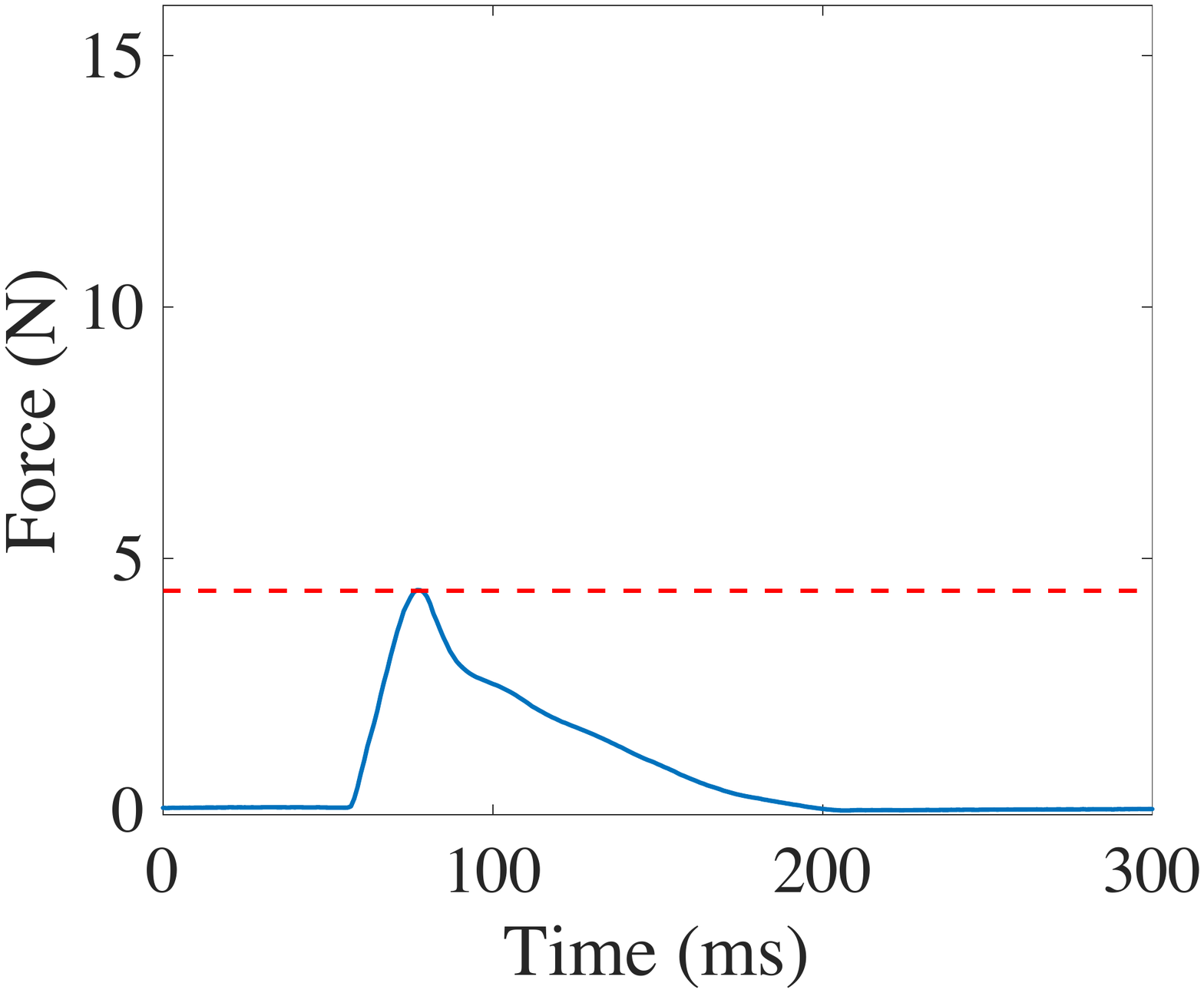}}}\\
        \subfloat[DFT of selected tap, vertical line shows spectral centroid.]{\includegraphics[clip, trim=0.cm 4.5cm 1.5cm 5.5cm,width=0.45\linewidth]{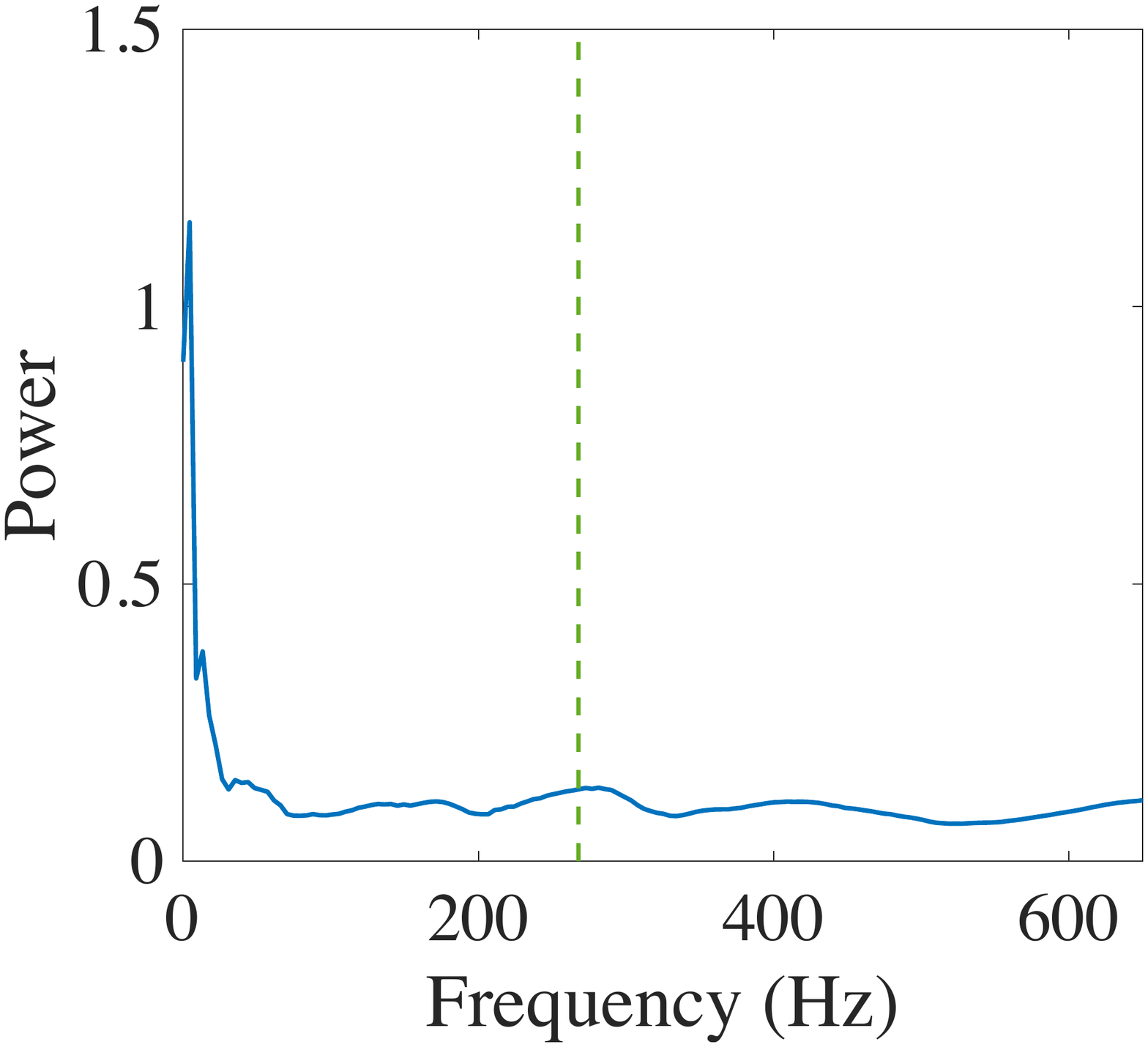}
        {\includegraphics[clip, trim=0.cm 4.5cm 1.5cm 5.5cm, width=0.45\linewidth]{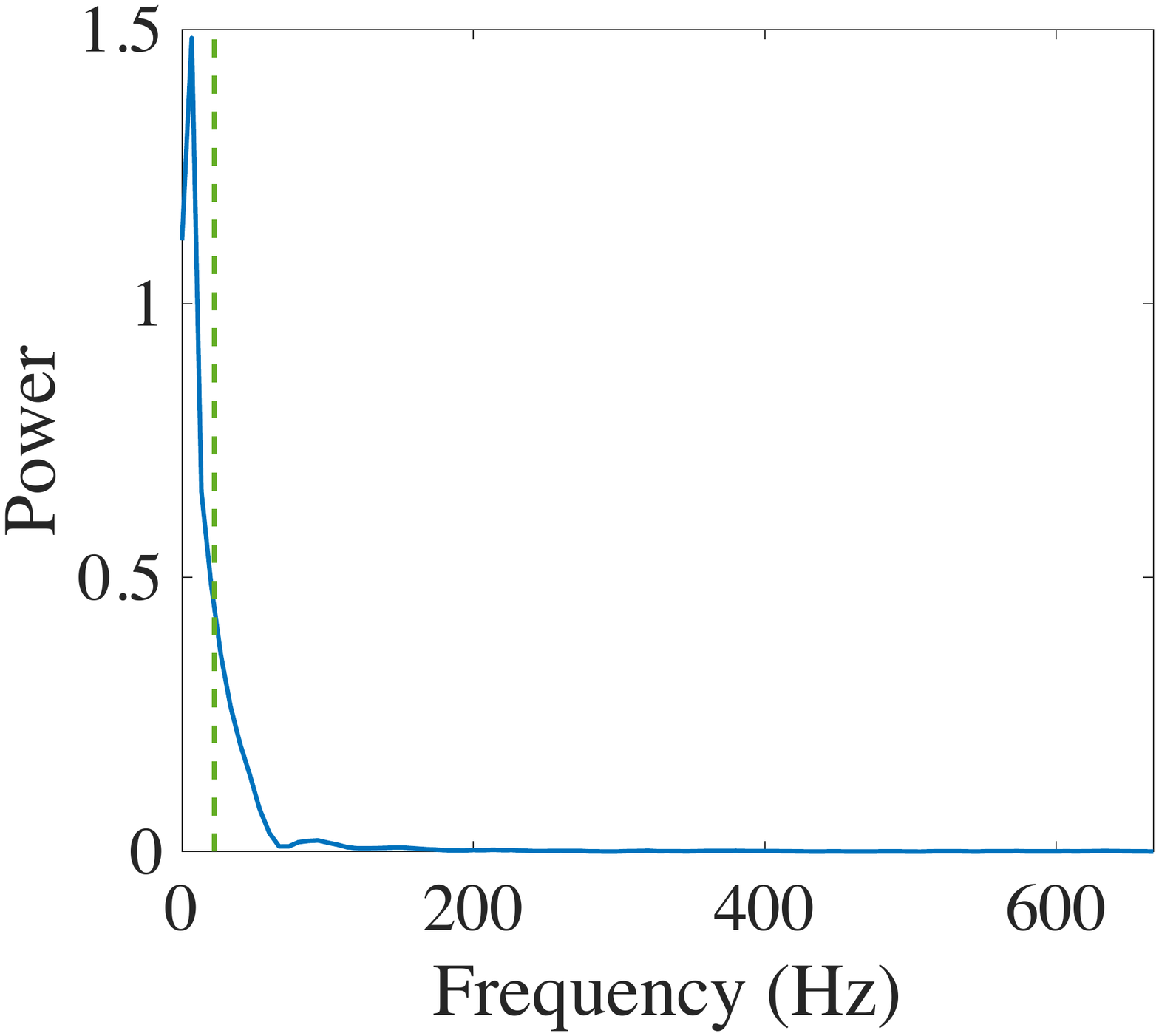}}}
        \caption{{Data processing steps for tapping data. (left) hardest plate, (right) softest plate.}}\label{fig:fft}
\end{figure}

Magnitude and frequency spectra are essential parameters in determining the perception of certain haptic properties of a surface, such as hardness. To understand the relative effects of stiffness and physical hardness in the perceptual hardness of the ETHD, we compare the force responses during the interaction between the stylus tip and end-effector plates.
The experimenter tapped 30 times on each plate for each stiffness {with a constant speed} and a tapping rate in the range of human tapping. {We varied the stiffness between the minimum and maximum stiffness (200~N/m and 2000~N/m) in increments of 100~N/m.} Figure~\ref{fig:fft} shows an example of how we processed the tapping data. {First we cropped the recorded tap signals by removing the first 3 seconds and after 10 seconds to avoid noise. Then we get an average from all the peaks of the taps in that 7 seconds range (red dashed line in Figure~\ref{fig:fft}-(a)), and select the tap that is closest to that line (Figure~\ref{fig:fft}-(b)).
We then calculated the Discrete Fourier transform (DFT) for each signal (Figure~\ref{fig:fft}-(c)).}
The spectral centroid $SC$ of the DFT was calculated for each signal to compare the frequency spectra of signals over time:

\begin{equation} 
SC=\frac{\sum_{n=0}^{N-1}f(n)x(n)}{\sum_{n=0}^{N-1}x(n)}
\end{equation}
where $X(n)$ is the amplitude of the DFT in the $n$th window at $f(n)$ frequency. This value is shown as the green vertical dashed line in Figure ~\ref{fig:fft}-(e),(f)).


\subsection{Result of Experiment 1}

\begin{figure}[tb]
  \centering
  \includegraphics[width=0.98\columnwidth]{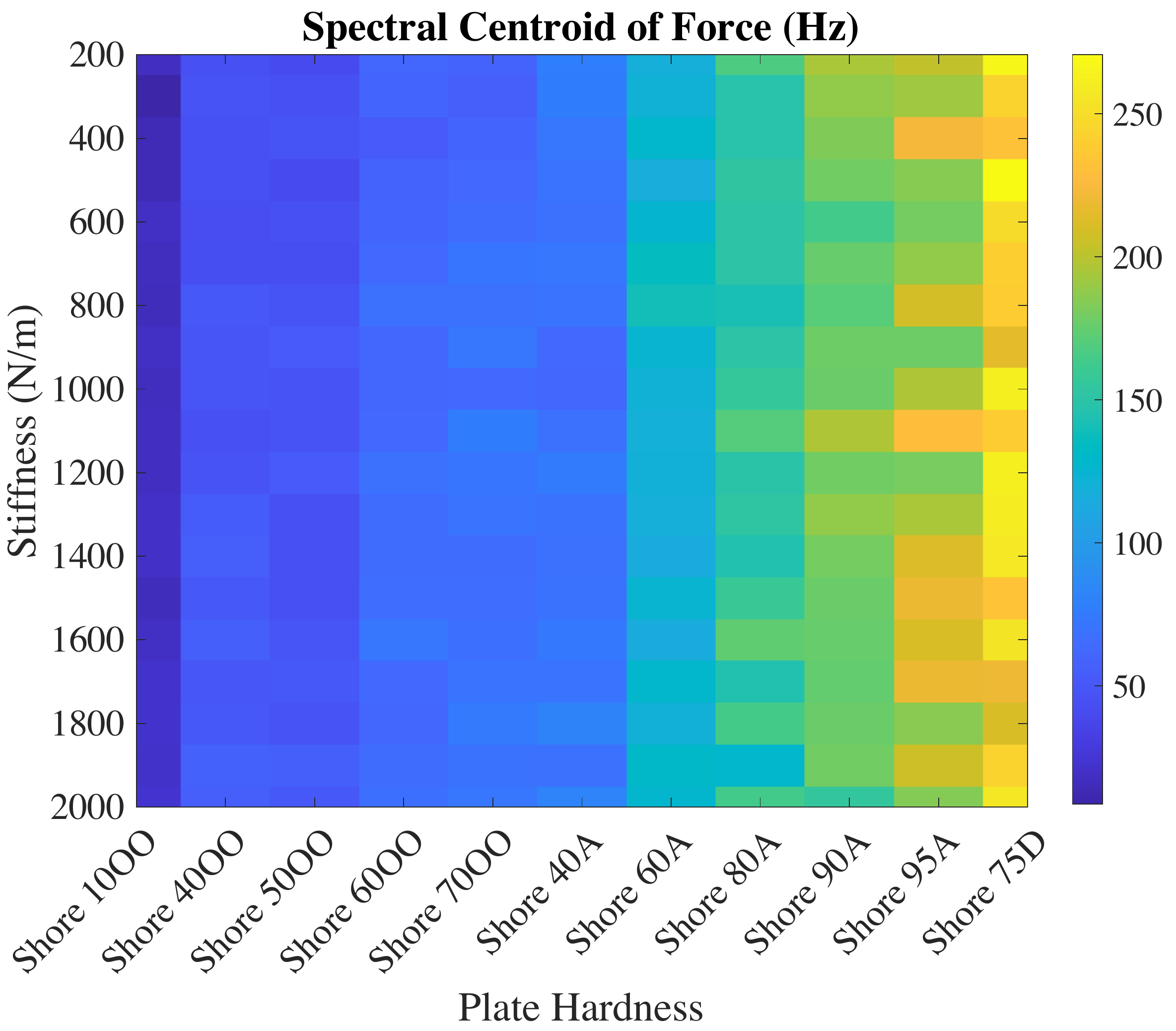}
  \caption{Measured spectral centroid of force for each hardness of plates versus device stiffness}\label{fig:centroid}
\end{figure}

Figure~\ref{fig:centroid} shows the spectral centroid of the tapping force for 11 plates of varying hardness under a range of rendered stiffness values.  
{A two-way ANOVA with hardness and stiffness as factors shows that spectral centroid of force is statistically significant across different hardnesses ($F(10,180)=1192.6, p<0.001$), but is not significant across different stiffnesses.}

\begin{figure}[tb]
  \centering
  \includegraphics[width=0.9\columnwidth]{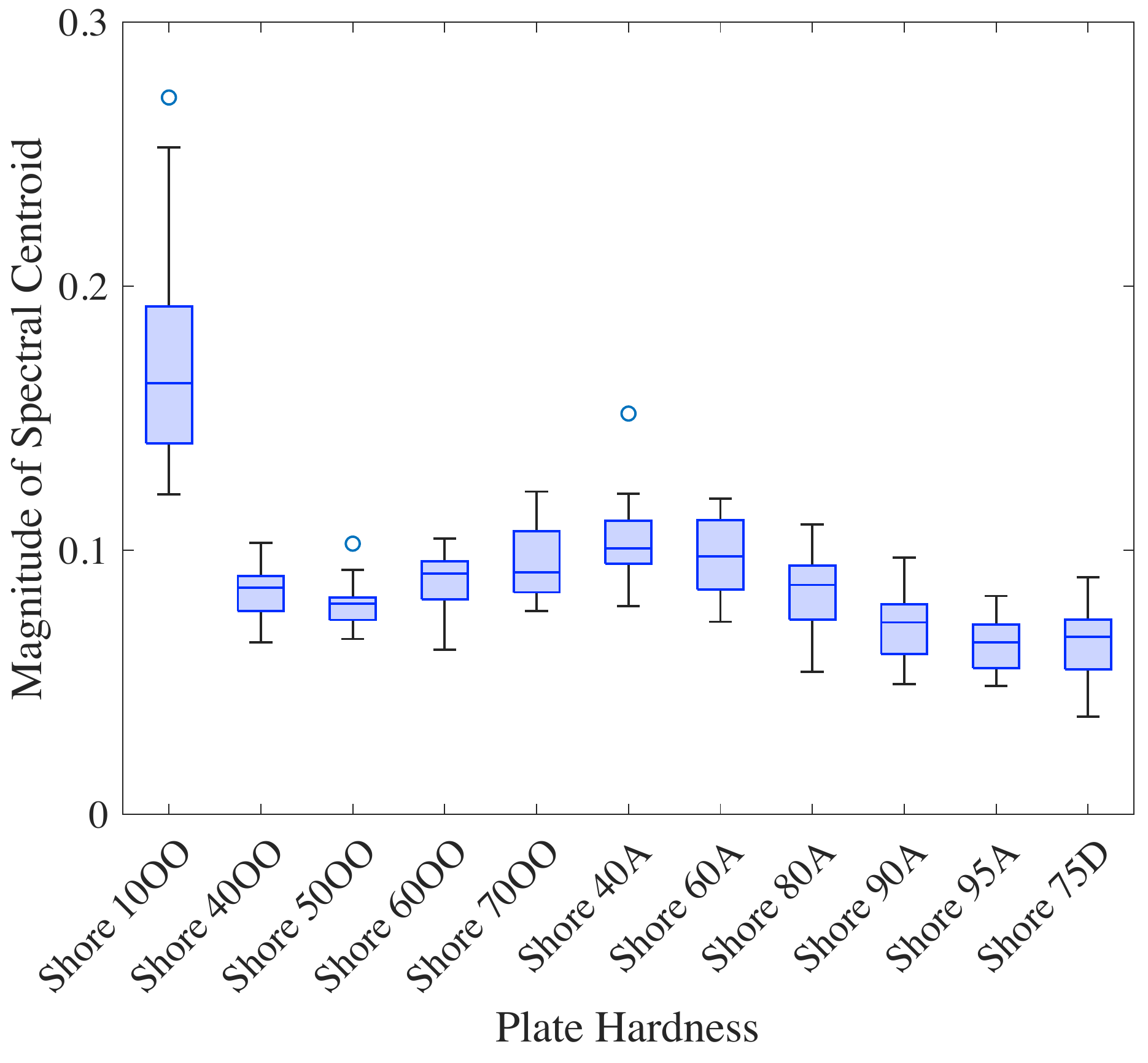}
  \caption{Magnitude of spectral centroid of force for each plate hardness.}\label{fig:power-centroid}
\end{figure}

Figure~\ref{fig:power-centroid} shows the {magnitude of the spectral centroid} of forces across all stiffnesses and hardnesses. 

\begin{figure}[tb]
  \centering
  \includegraphics[width=0.9\columnwidth]{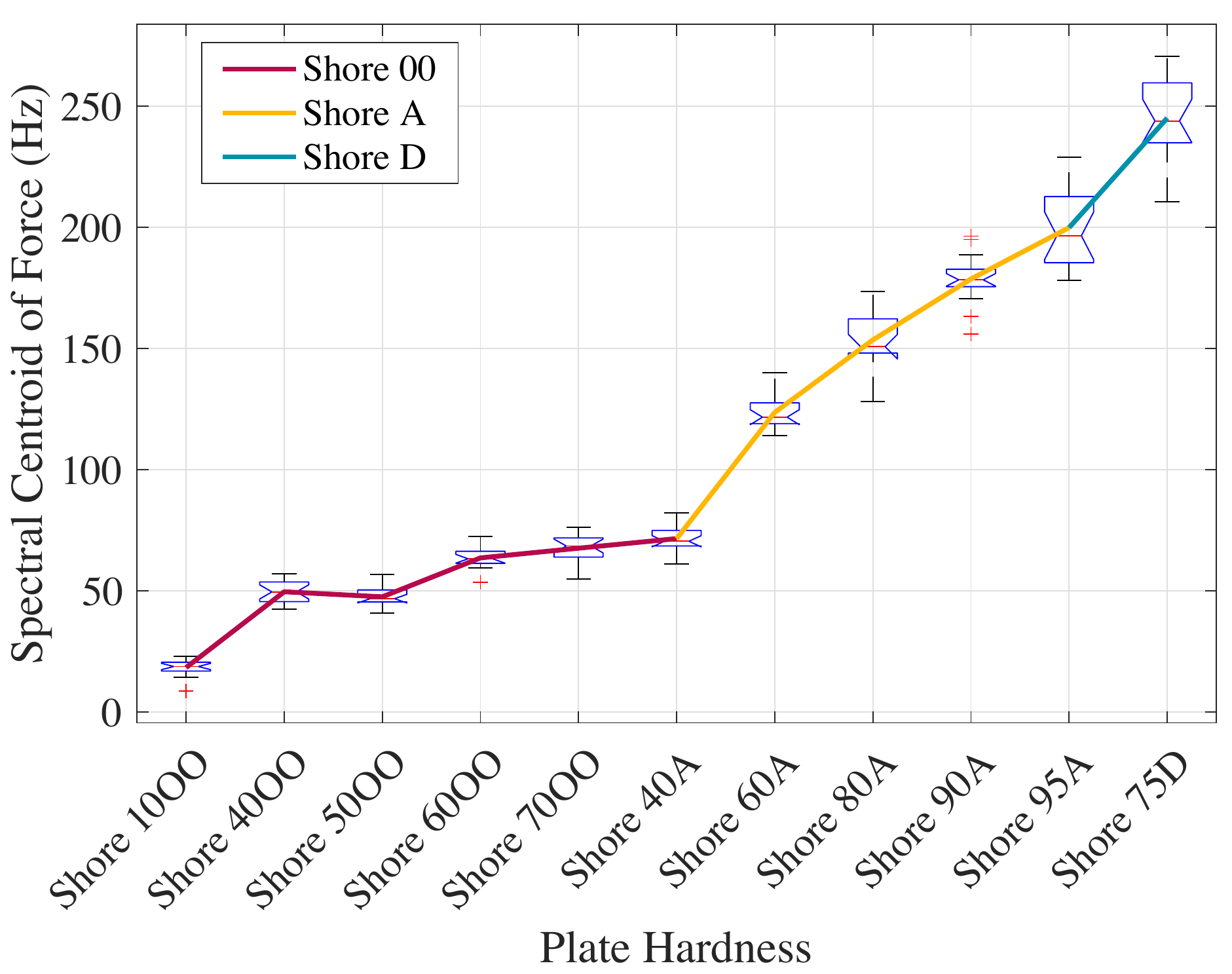}
  \caption{Average of spectral centroid of force for every stiffness and hardness}\label{fig:shore}
\end{figure}

To better understand the differences between the different shore durometer of plates, Figure \ref{fig:shore} shows the boxplot of spectral centroid of force versus the hardness of plates, with a connected line indicating their averages, and their colors show the different categories of shore durometers.

\begin{figure}[tb]
  \centering
  \includegraphics[clip, trim=0cm 0cm 4.95cm 0cm,width=\columnwidth]{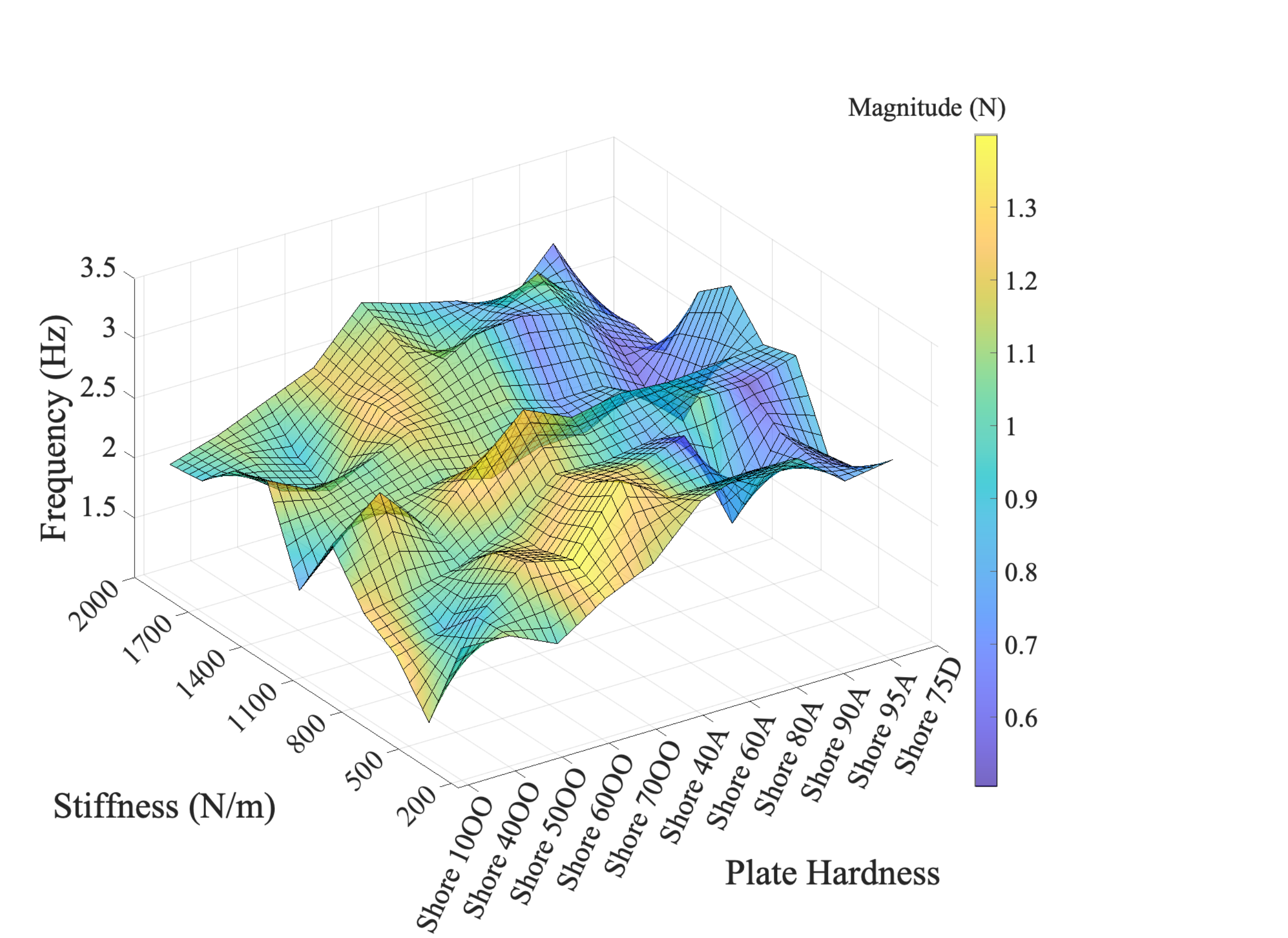}
  \caption{{\naghme{Magnitude of dominant frequency of forces versus plates hardness and stiffness.}}}\label{fig:freqpow}
\end{figure}

For each hardness and stiffness, we also found the dominant frequency of the tapping force (i.e., the frequency at which the maximum magnitude in the DFT happens). The results are shown in Figure~\ref{fig:freqpow}.
{A two-way ANOVA shows that dominant frequency of force is statistically significant across different hardnesses ($F(10,180)=14, p<0.001$) and across different stiffnesses ($F(18,180)=3.22, p<0.001$).
A second two-way ANOVA shows that the magnitude of dominant frequency is also statistically significant across different hardnesses ($F(10,180)=32.96, p<0.001$) and across different stiffnesses ($F(10,180)=3.6, p<0.001$).}

\subsection{Discussion of Experiment 1}
{Figure~\ref{fig:centroid} shows that the spectral centroid of force only changes based on the hardness of the plates and is not affected by the rendered stiffness. In this experiment we tested 11 plates, which covered the range from extra soft to extra hard in Shore durometer scale. Figure~\ref{fig:shore} indicates that the varitation in spectral centroid of force across different plate hardnesses is highly dependent on the materials' Shore category.

To shorten the next experiment, we used these results to reduce the materials tested. We divided the minimum and maximum spectral centroid of force into five levels, and found the hardness in Figure~\ref{fig:shore} that was closest to each division. This resulted in a set of five materials of different hardnesses, one in each of the categories extra soft, soft, medium, hard, and extra hard.} These materials are listed in Table~\ref{table:short-hardness} and shown in Figure~\ref{fig:tip}.


\subsection{Experiment 2}
\begin{figure}[tb]
  \centering
  \includegraphics[width=1\linewidth]{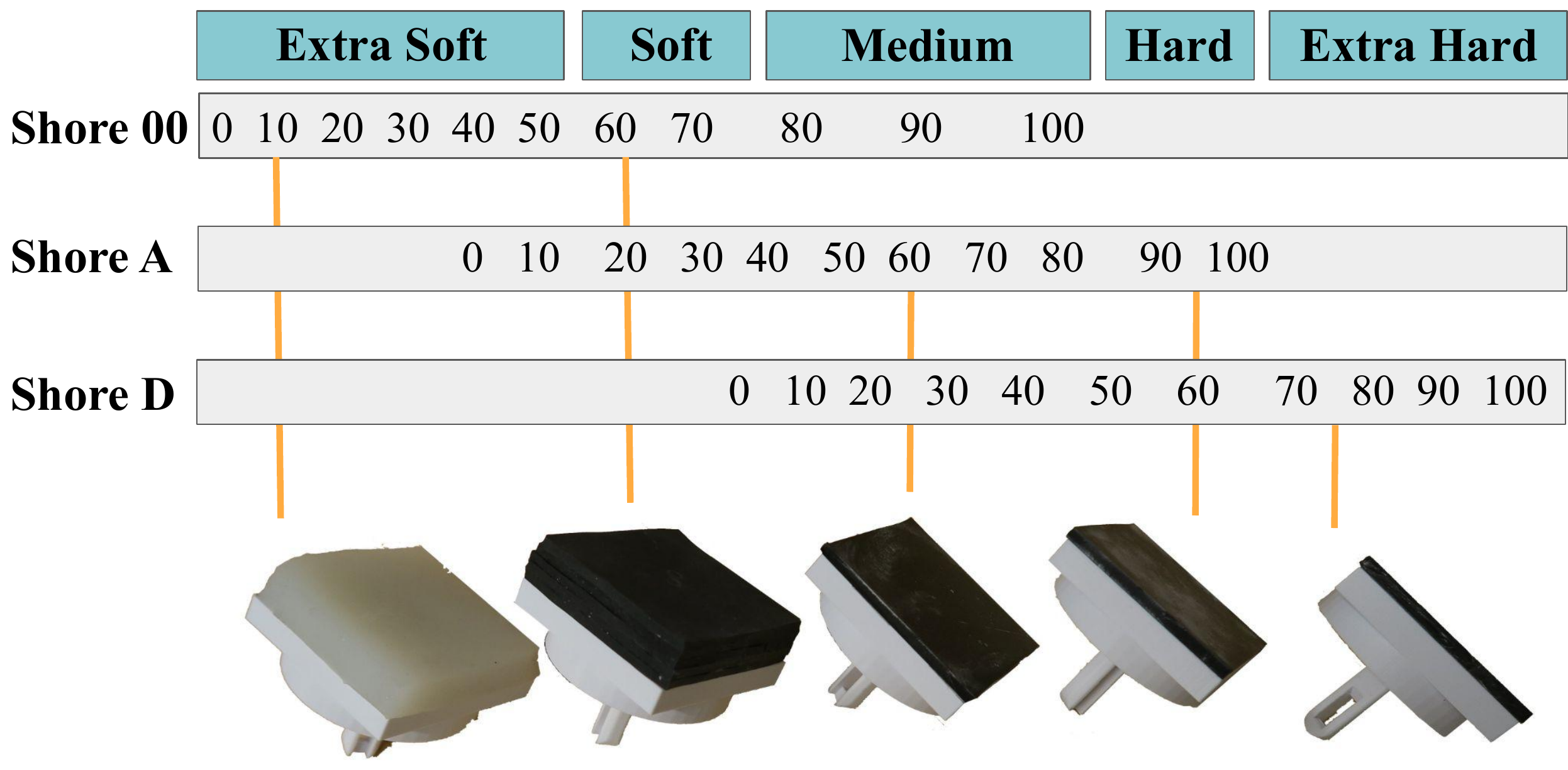}
  \caption{{Selected hardnesses based on the durometer standard table. The plates with different hardnesses are attached to a 3d printed mount for connection to the haptic device.}}\label{fig:tip}
\end{figure}

In this experiment, we tested the participants' ability to distinguish between different rendered stiffnesses when interacting with a constant hardness surface attached to the device's end-effector. For each plate, we determined the Weber fraction for stiffness, {which is the ratio of the just noticeable difference to the intensity of the stimulus}, by having the participant compare their perception of two rendered stiffnesses at a time through tapping. 
As discussed above, we tested a set of five plates, one from each categories of extra soft, soft, medium, hard, and extra hard (Figure~\ref{fig:tip}, Table~\ref{table:short-hardness}).
\vspace{-0.2in}
\begin{center}
\begin{table}
\caption{{Selected plates in different hardness categories}}\label{table:short-hardness}
\begin{tabular}{ |c|c|c| } 
\hline
Plate No. & Shore No. & Range \\ 
\hline
\hline
P1 & Shore 10 OO & Extra Soft \\ 
\hline
P2 & Shore 60 OO & Soft \\ 
\hline
P3 & Shore 60 A & Medium  Soft \\ 
\hline
P4 & Shore 90 A & Hard \\ 
\hline
P5 & Shore 75 D & Extra Hard \\ 

\hline
\end{tabular}
\end{table}
\end{center}

\naghme{We conducted a pilot study with four participants to select the set of reference stiffnesses to test. In this pilot study, we used a set of eight reference stiffnesses from 500~N/m to 2250~N/m in increments of 250~N/m. We found the Weber Fraction showed a monotonic linear trend with stiffness, indicating that is was not necessary to test all 8 reference stiffnesses in the full experiment. Therefore, we selected a set of four stiffnesses from 500~N/m to 2000~N/m with an increment of 500~N/m in order to shorten the experiment and minimize fatigue.}



Participants followed a Two Alternative Forced Choice (2AFC) procedure by tapping on the end-effector plate with the reference stiffness and test stiffness {in randomized order} and responded to the prompt: "Which surface feels harder?"
They controlled the stimuli and responded to the prompt using four labeled keys on a keyboard. Two keys toggled back and forth between the test and reference stiffnesses, and the other two keys were used to respond to the prompt by selecting the stimuli (1 or 2) that they perceived to be harder.


For each reference stiffness, the test stiffness was chosen following a modified staircase method, as explained below. This psychophysics method was integrated with Chai3d to control the haptic loop of the device.



\subsubsection{Modified Staircase method}
We used an adapted staircase method to select the test stiffness based on the participant's previous responses. For each plate and each reference stiffness, the test stiffness began at the minimum stiffness ($200~N/m$) and varied with the related step size. The trial terminates after five reversals.
{However, in our pilot study we found that if there would be a reversal by mistake at early steps, the average threshold will be much lower than the actual threshold because the final threshold is the average between all reversals.}

After uncovering this issue in our pilot study, we decided to implement an additional rule we found to be helpful in the accuracy of detecting the threshold by ignoring responses that are likely mistakes. After each wrong answer, if the participant can answer correctly for the next four times, we ignore the last change of direction. The reason is that the probability of answering correctly four times in a row by chance is only $6.25\%$, which is very low. Therefore, it is likely that the previous change of direction was a mistake, caused by any reason.  {These mistakes could happen because of change in tapping speed, maintaining contact with surface, sudden change of body position, or other distractions.}

In a pilot study, we tested this staircase method both with a step size of 50~N/m and 100~N/m. We found that the Weber fractions for reference stiffnesses higher than 500~N/m were similar for both 50~N/m and 100~N/m. Therefore, in the full study, we used a step size of 50~N/m for the reference stiffness 500~N/m, and a step size of 100~N/m for all other reference stiffnesses. This variation in step size was chosen to limit the experiment length and avoid muscle fatigue, which would affect the accuracy of the results.

\begin{figure}
        {\includegraphics[clip, trim=0.5cm 1.1cm 1.5cm 1cm,width=0.87\columnwidth]{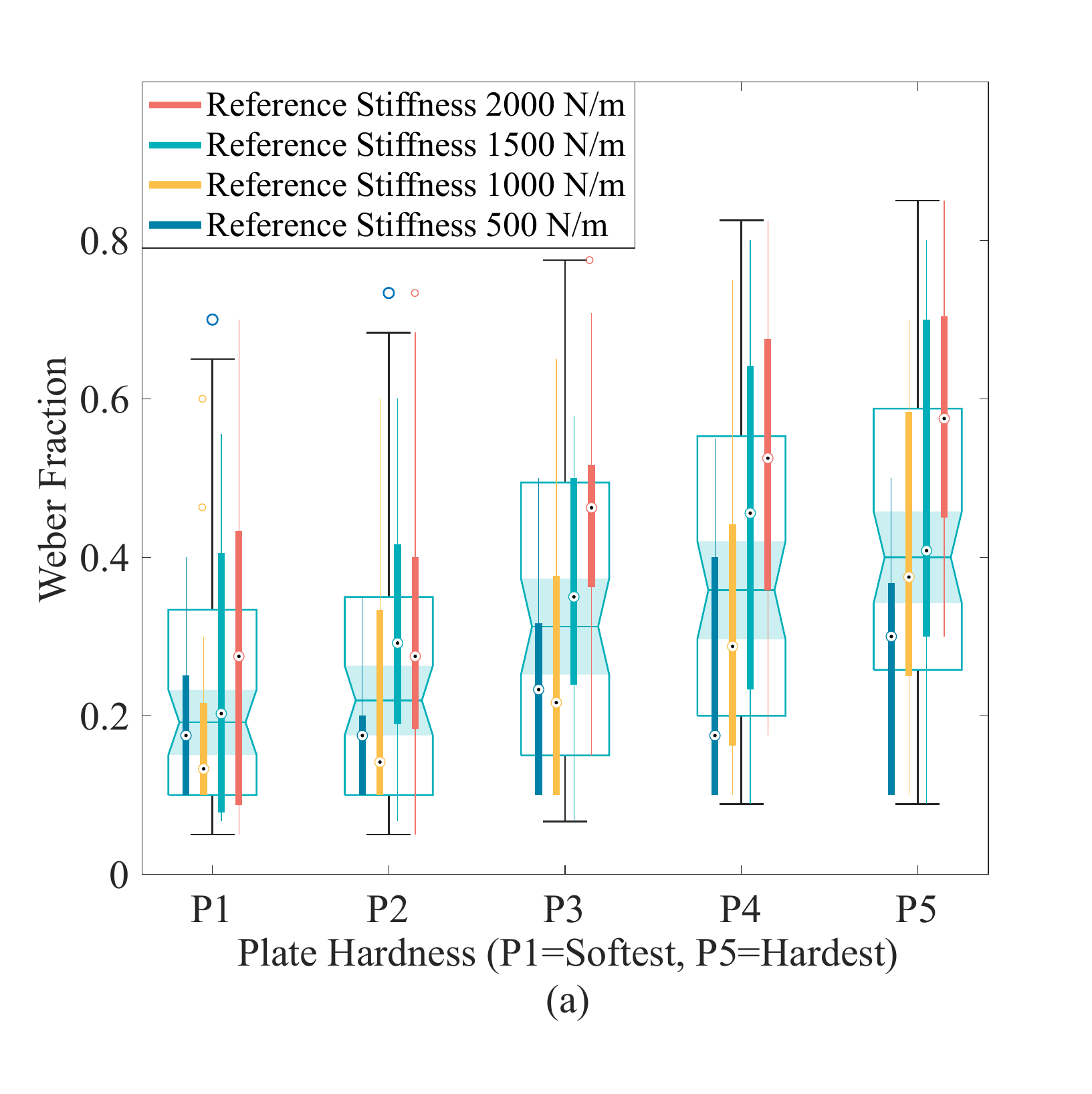}}\\
        {\includegraphics[clip, trim=0.5cm 2.4cm 1.5cm 1.1cm,width=0.87\columnwidth]{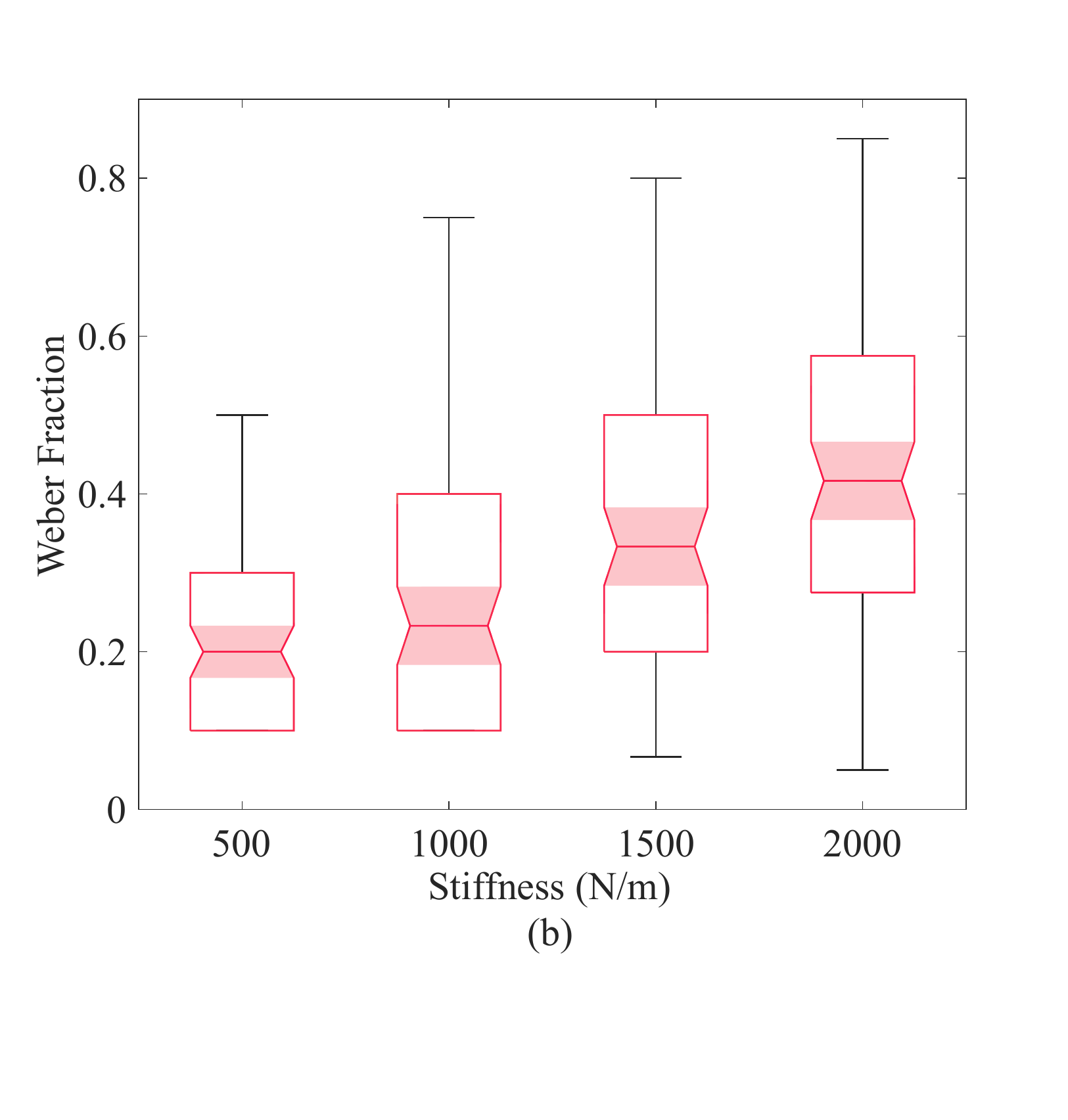}}\\
        {\includegraphics[clip, trim=0.5cm 0cm 1.5cm 2cm,width=0.87\columnwidth]{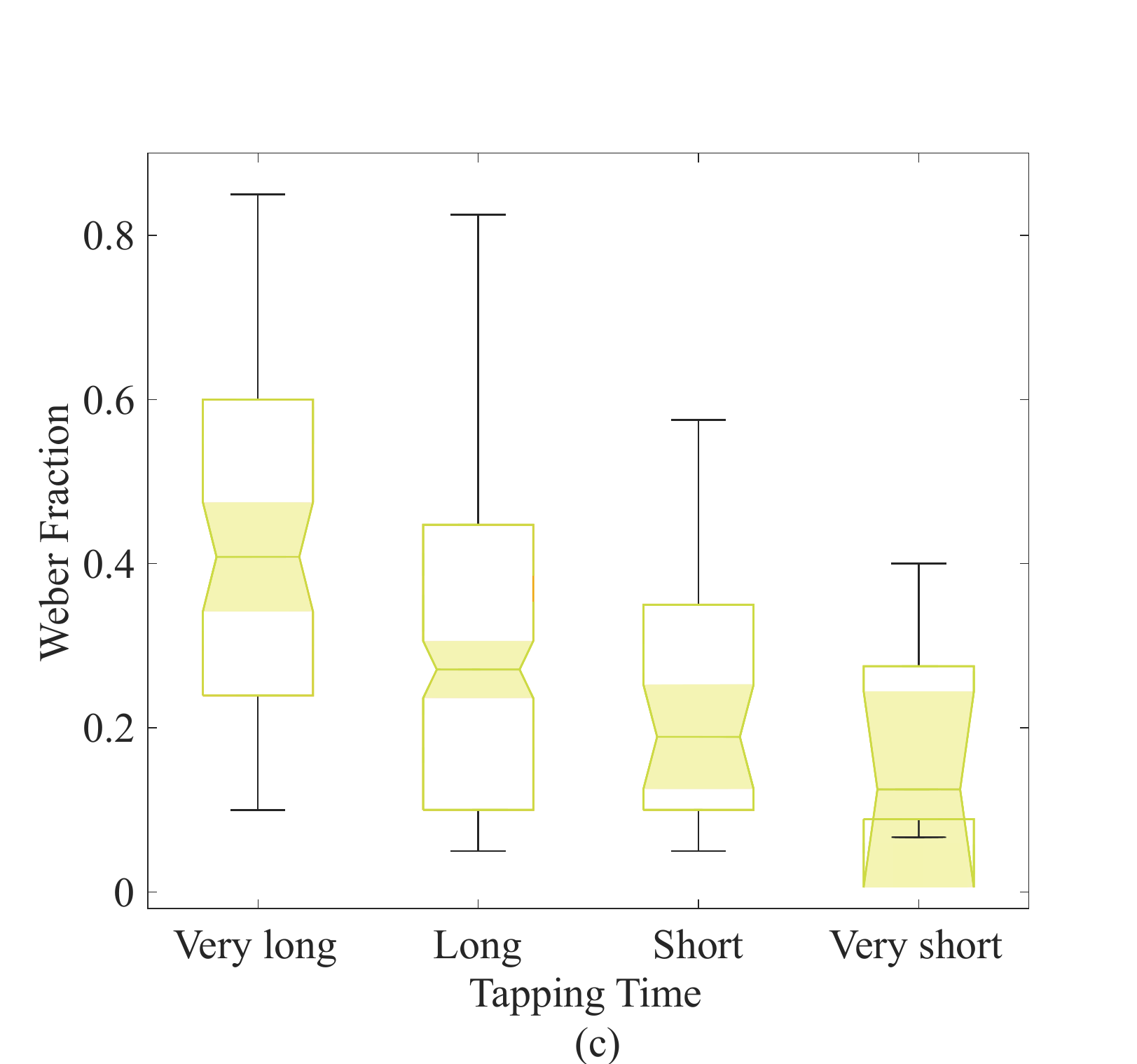}}\\
        \vspace{-0.1in}
       \caption{{(a) Weber fraction of stiffness versus hardness of plates. Within each plate, smaller set of boxplots indicates how weber fraction of stiffness varies among different reference stiffnesses. (b) Weber fractions of stiffnesses versus test stiffnesses, and (c) versus tapping times.} \label{fig:wf}}
\end{figure}

\subsubsection{User Study}
We recruited 20 participants (20-35 years old; 8 female, 12 male; one left-handed).
They wore headphones playing white noise to block any audio cues from the device. They sat on a chair with the device located next to them on a table, and held the stylus in their dominant hand. A keyboard was placed in front of them with four keys labeled for switching between stiffnesses and responding to the prompt, which they controlled using their non-dominant hand.
When the experiment started, participants were instructed to not look at the device during tapping, and to focus on a stationary ball on the monitor in front of them that changes color when the stylus force sensor detects a tap. They were allowed to look at the device and adjust the stylus' location between taps.

Participants were instructed to tap on each stimulus for about 10 seconds while keeping the stylus vertical. We did not limit how they hold the tool to have a variety of grasping styles, since different contact location of stylus with hand influences the perceptual threshold of  tactile information~\cite{zamani19}. They were allowed to switch back and forth between the two stiffnesses using the keyboard as many times as they wanted before giving their response.

\subsection{Results of Experiment 2}
Figure~\ref{fig:wf}-(a) shows the boxplot of Weber fraction for all reference stiffnesses versus the hardness of plates (large boxplots). To understand and visualize the effect of reference stiffness, inside each large box we show a different boxplot that groups the results based on their reference stiffness.


A two-way ANOVA on the Weber fraction for all stiffnesses with plate hardness and reference stiffnesses ($500~N/m, 1000~N/m, 1500~N/m,$ and $ 2000~N/m$) as the factors indicated that the Weber fractions were statistically different across different hardnesses ($F(4, 395) = 18.58, p<0.001$), and across different reference stiffnesses ($F(4, 396) =28.63, p<0.001$).

A Tukey’s post-hoc pairwise comparison test further evaluated the effects of the plate hardness on the Weber fraction of stiffnesses. The results showed significant differences between the Weber fractions of all plates except for the ones closest in hardness. For example, P1 was significantly different from P3, P4, and P5, but not from P2.

We ran a set of one-way ANOVAs on the Weber fraction of stiffness within each plate hardness with reference stiffness (500~N/m, 1000~N/m, 1500~N/m, 2000~N/m) as factor. The Weber fraction was significant across the reference stiffnesses for P2 ($F(3,96)=4.18, p=0.009$), P3 ($F(3,96)=7.55, p<0.001$), P4 ($F(3,96)=7.39, p<0.001$), and P5 ($F(3,96)=8.83, p<0.001$). The Weber fraction did not significantly vary across the reference stiffnesses for P1 ($F(3,96)=2.27, p=0.087$). 
A Tukey’s post-hoc pairwise comparison test further evaluated the effects of the reference stiffness on the Weber fraction of stiffnesses within each plate hardness. The results show the following:
{For P2, the results showed significant differences between the Weber fractions of reference stiffness 500~N/m with both 1500~N/m and 2000~N/m.
For P3, the results showed significant differences between the Weber fractions of reference stiffnesses 500~N/m and 1000~N/m with 2000~N/m.
For P4 and P5, the results showed significant differences between the Weber fractions of reference stiffness except for the ones closest together. For example, 500~N/m was significantly different from 1500~N/m, and 2000~N/m, but was not different from 1000~N/m.}

Figure~\ref{fig:wf}-(b) shows a boxplot of the Weber fraction of stiffnesses versus the reference stiffnesses combined for all hardnesses. 
We ran a one-way ANOVA on this combined Weber fraction of stiffness with reference stiffness as factor. This analysis indicated a statistically significant difference in Weber fraction across reference stiffnesses ($F(3,396)=24.31, p<0.001$).
A Tukey’s post-hoc pairwise comparison test further evaluated the effects of the reference stiffness on the Weber fraction of stiffnesses. The results showed significant differences between the Weber fractions of reference stiffness except for the ones closest together. For example, 500~N/m was significantly different from 1500~N/m, and 2000~N/m, but was not different from 1000~N/m.

Figure~\ref{fig:wf}-(c) shows the boxplot of Weber fraction of stiffnesses versus tapping time.
Tapping time is the duration of one average tap for each trial (shown in Figures~\ref{fig:fft}-(b). {The minimum and maximum of all of the selected tap durations were 20~ms, and 288~ms respectively. We grouped these durations to four categories: ``Very long" (221~ms-288~ms), ``Long" (154~ms-221~ms), ``Short" (87~ms-154~ms), and ``Very short" (20~ms-87~ms).} We ran a one-way ANOVA on the Weber fraction of stiffness with tapping time as factor. This analysis showed that the Weber fraction is statistically significant across different tapping times ($F(3,396)=10.55, p<0.001$). 
{A Tukey’s post-hoc pairwise comparison test further evaluated the effects of the tapping times on the Weber fraction of stiffnesses. The results showed significant differences between the very short duration taps with taps of short, long, and very long duration.}

{We calculated the speed of the taps by taking the derivative of the position sensor data, and measured the force of the taps using the Z-axis data of the force sensor.
Then }we ran a 4-way ANOVA on the Weber fraction for each reference stiffness with plate hardness, participant, force, and speed as factors. \naghme{In order to include force and speed as factors we converted them to categorial variables.} The ANOVA for reference stiffness 500~N/m found that the Weber fraction was statistically significant for hardness ($F(4,74)=3.69, p=0.009$) and participant ($F(19,74)=3.39, p<0.001$). For 1000~N/m, there was a significant difference in Weber fraction for hardness ($F(4,74)=11.8, p<0.001$), participant ($F(19,74)=5.99, p<0.001$), and force ($F(1,74)=9.62, p=0.003$). For $1500~N/m$, there was a significant difference in Weber fraction for hardness ($F(4,74)=16.46, p<0.001$), participant ($F(19,74)=11.99, p<0.001$), and force ($F(1,74)=9.79, p=0.003$). For 2000~N/m, there was a significant difference in Weber fraction for hardness ($F(4,74)=26.35, p<0.001$) and participant ($F(19,74)=14.85, p<0.001$). No other factors were found to be significant.

\begin{figure}[tb]
  \centering
  \includegraphics[clip, trim=0.0cm 0cm 0cm 0cm,width=\linewidth]{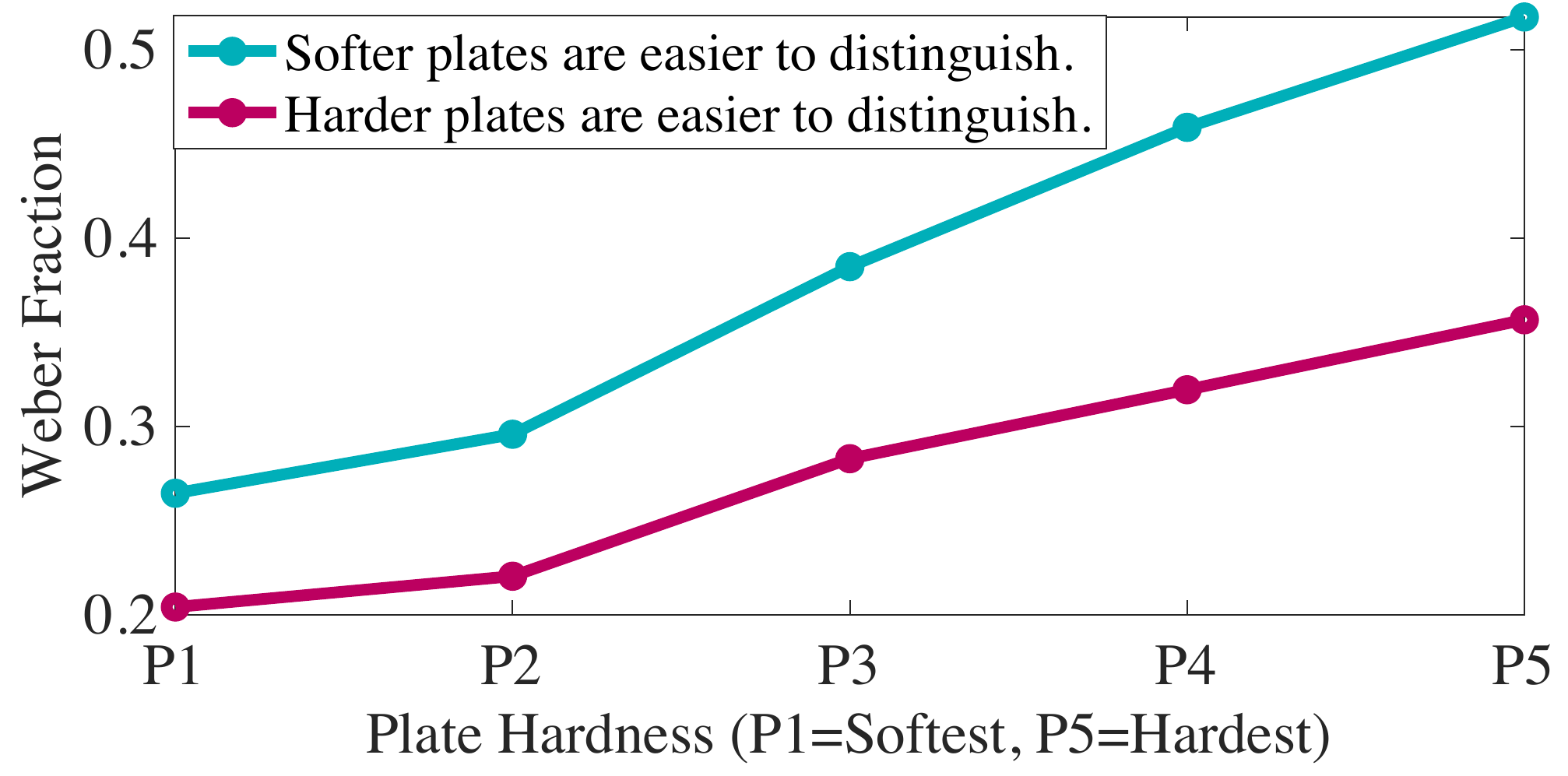}
   \caption{Comparison of Weber fraction of stiffnesses versus plate hardness for people who think hard surfaces are easier to distinguish versus people who think soft surfaces are easier. \naghme{Each point indicates the average Weber fraction for all four reference stiffnesses (500~N/m, 1000~N/m, 1500~N/m, and 2000~N/m) for that plate hardness.}\vspace{-0.3in}}\label{fig:wf-easy}
   
\end{figure}

In a post-experiment survey we asked the participants to indicate for which type surface, hard or soft, they thought it was easier to distinguish the difference in stiffness. Nine participants wrote that the found hard surfaces easier to distinguish, and 11 participants wrote that they found softer surfaces easier. Note that almost all participants could distinguish the stiffnesses for softer surfaces better, as indicated by a smaller Weber fraction, even if they thought harder surfaces are easier to distinguish. Figure~\ref{fig:wf-easy} shows this result.  

\section{Discussion}

{We proposed a new approach for rendering both hard and soft surfaces that overcomes limitations of existing impedance and admittance-type haptic devices. In our approach we modified an impedance haptic device by detaching the stylus from the body of the device and using the end-effector as the primary interaction point, creating an ETHD.
Our design has two significant benefits; the system can generate a hard surface even given stability and stiffness constraints of the impedance device, and the user can move the stylus effortlessly while there is no contact due to its untethered nature.


This system rendered both hardness and stiffness of a virtual object. The studies conducted in this paper focused on determining if the rendered hardness could mask the rendered stiffness, allowing users to perceive a hard surface even when the underlying stiffness was low. This masking effect is a key advantage of our system because it allows us to still render a perceptually hard surface even if the device is limited in its stiffness output. Therefore, in this study, we focus on a low-cost haptic device, the Novint Falcon, that is not capable of high force output.
We believe that if the concept works with a low-cost device, then we can get even a better result if we use a more capable haptic device. 

We evaluated the hardness perception of tapping on the end-effector of the ETHD, while both end-effector hardness and device stiffness were changed. Our studies provided important insights
into the strengths of our rendering approach and the importance of the masking effects between stiffness and hardness.

As discussed in Experiment 1 and based on~\cite{Culbertson17}, a good predictor of  hardness perception ratings is the spectral centroid of the tapping transient. The tapping transient has a higher spectral centroid for harder surfaces and a lower spectral centroid for softer surfaces as shown in Figure~\ref{fig:fft}. Interestingly, we did not see a noticeable difference in the spectral centroid for a given hardness when the rendered stiffness was changed. This indicates that the transient portion of the force during contact was dominated by the hardness of the surface, although the stiffness may have affected the lower frequency or steady-state forces of the contact. This quantitative result suggests that the hardness perception might not be greatly affected by underlying rendered stiffness. However, the dominant frequency of force does change with both hardness and stiffness, indicating that the rendered stiffness does affect the frequency of the tapping transients. The magnitude of the spectral centroid was fairly constant across the 11 tested materials except for the softest material (Shore 10OO), which showed a significantly higher magnitude. One possible reason for this \naghme{is that the force is overdamped during the tap on the softest plate (Figure~\ref{fig:fft} shows an example of this signal behaviour.}
We qualitatively tested the effect of the stiffness on perceived hardness in Experiment 2.

In this experiment, we found that the Weber fraction of stiffness is highly dependent on the hardness of the end-effector. The harder the plate is, the higher the Weber fraction is. 
These results show that when interacting with a hard surface, it is difficult for users to recognize and understand variations in forces from the device due to a change in stiffness. The ability of a user to distinguish changes in stiffness becomes worse as the hardness increases. As such, the hardness of the end-effector acts to mask the stiffness of the device. \naghme{Similar masking effects on other haptic modalities have also been found~\cite{rank,zook2021effect}.}
This result shows that as long as we augment the interaction point of the end-effector with a small hard material in our approach, it is not necessary for the device to generate the actual desired force because the perceived hardness increases without actually increasing the stiffness. For example if the user cannot distinguish between 1500~N/m and 2000~N/m, there is no need to use a device with better actuators to generate 2000~N/m. 



Our results also show that the Weber fraction increases with increased reference stiffness, regardless of the hardness. This means that users cannot easily distinguish between changes in rendered stiffness when the underlying stiffness is high. This also shows promise for our system being used with low-cost haptic devices that cannot generate high stiffnesses or do not have accurate stiffness control. 

\naghme{As shown in Figures~\ref{fig:wf-easy} and \ref{fig:wf}-(a), on average the Weber fraction of stiffnesses are lower when participants tapped on softer plates. 
Figure~\ref{fig:wf-easy} compares the Weber fraction of the participants result based on their answer of which plates are easier to distinguish. Subjects who chose softer plates as easier surfaces to distinguish always had higher Weber fraction, meaning they had lower perception of the underlying stiffness compared to the other group. Both groups regardless of which plates they choose as easier to distinguish, always distinguished the soft plates better.}

The analysis shows that additional factors, including how the user interacts with the device, may also affect their perception of the hardness during tapping. For example, the more quickly a user taps (i.e., the less time they stay in contact with the end-effector), the more difficult it is for them to understand changes in stiffness. This difference is important because users will likely stay in contact with the surface for a longer time during interactions with virtual objects. We leave it for future research to determine how the masking effects seen in tapping translate to static and sliding contact. 
We also saw an effect of force during the tap on the Weber fraction for reference stiffnesses 1000~N/m and 1500~N/m, and users that applied a larger force to the end-effector could more easily distinguish between the rendered stiffnesses. However, we did not see an effect of tapping speed on Weber fraction in our experiment, meaning that the hardness perception and masking effect was not affected by the speed at which the user contacted the end-effector. 


These results show that our new system helps to render virtual surfaces more realistically and with a lower cost over traditional rendering methods through masking effects. Our system has the added benefit of minimising friction and inertia of the interaction through its untethered design. 
Our system can be used effectively in many applications where rendering hard surfaces is required.
For example, most dental schools currently use manikins for training, which are not realistic. The only commercialized haptic dental simulator available in some schools (Moog Simodont Dental Trainer) uses an impedance-type haptic device that cannot generate very hard surfaces. Similarly, in orthopedic training they mostly use PVC pipes to simulate bone drilling because there is no haptic device that can simulate both free space and hard surfaces. We believe this haptic system could enable a shift in current education training systems in schools. This system could also play a significant role in different teleoperation applications because it overcomes previous limitations, including generating high and low force-feedback and tactile feedback at the same time. 
}

\section{Conclusions and Future Work}
A major problem in many haptic systems is the rendering of hard objects. This is an issue that admittance devices cannot fix because they have high friction and inertia, which negatively affects rendering in free space. In applications like dental training, we usually need to simultaneously render free space and dense space, for example adjusting the stylus before making contact with the tissue. The device should provide zero friction in free space with a sudden high force/friction when it contacts the tissue. This problem has not been fully solved since the stylus is connected to the end-effector, and we have to carry the weight and friction of joints even when we do not want to provide any force.
In this paper we have suggested solving this problem by using augmented reality methods and untethered haptic tooling. We attached plates with different hardnesses on the end-effector and created a stylus with embedded sensors to track location and force during contact.
We conducted two experiments to determine how changing the hardness can affect the perception regarding the device's stiffness.

The results show us the importance of hardness in masking the stiffness of the haptic device and indicate that more attention should be paid towards rendering hardness when designing haptic systems.
This paper compares the perceived stiffness while the hardness is constant. In future work, we will further evaluate this system by determining the threshold for noticeable changes in hardness when the rendered stiffness is constant. We will also evaluate the realism of our rendering method by comparing real objects to virtual objects. 


\begin{thebibliography}{10}
\providecommand{\url}[1]{#1}
\csname url@samestyle\endcsname
\providecommand{\newblock}{\relax}
\providecommand{\bibinfo}[2]{#2}
\providecommand{\BIBentrySTDinterwordspacing}{\spaceskip=0pt\relax}
\providecommand{\BIBentryALTinterwordstretchfactor}{4}
\providecommand{\BIBentryALTinterwordspacing}{\spaceskip=\fontdimen2\font plus
\BIBentryALTinterwordstretchfactor\fontdimen3\font minus
  \fontdimen4\font\relax}
\providecommand{\BIBforeignlanguage}[2]{{%
\expandafter\ifx\csname l@#1\endcsname\relax
\typeout{** WARNING: IEEEtran.bst: No hyphenation pattern has been}%
\typeout{** loaded for the language `#1'. Using the pattern for}%
\typeout{** the default language instead.}%
\else
\language=\csname l@#1\endcsname
\fi
#2}}
\providecommand{\BIBdecl}{\relax}
\BIBdecl

\bibitem{CDCA}
{The Commission on Dental Competancy Assessments}. (2019) Dental educators
  conference report.

\bibitem{loomis1986tactual}
J.~M. Loomis and S.~J. Lederman, ``Tactual perception,'' \emph{Handbook of
  Perception and Human Performances}, vol.~2, no.~2, p.~2, 1986.

\bibitem{anderson1}
R.~Anderson, J.~Arro, C.~S. Hansen, and S.~Serafin, ``Audio-visual perception -
  the perception of object material in a virtual environment,'' in
  \emph{Augmented Reality, Virtual Reality, and Computer Graphics}, 2016, pp.
  162--171.

\bibitem{FLEMING201462}
R.~W. Fleming, ``Visual perception of materials and their properties,''
  \emph{Vision Research}, vol.~94, pp. 62--75, 2014.

\bibitem{gali2018technology}
S.~Gali and A.~Patil, ``The technology of haptics in dental education,''
  \emph{Journal of Dental \& Orofacial Research}, vol.~14, no.~2, 2018.

\bibitem{zamani21}
N.~Zamani, A.~Pourkand, H.~Culbertson, and D.~Grow, ``Plate-and-cable (pac)
  haptic device for orthopaedic training,'' in \emph{ISMR 2021- International
  Symposium on Medical Robotics}, 2021.

\bibitem{pourkand2017mechanical}
A.~Pourkand, N.~Zamani, and D.~Grow, ``Mechanical model of orthopaedic drilling
  for augmented-haptics-based training,'' \emph{Computers in biology and
  medicine}, vol.~89, pp. 256--263, 2017.

\bibitem{van}
F.~E. {van Beek}, D.~J.~F. {Heck}, H.~{Nijmeijer}, W.~M. {Bergmann Tiest}, and
  A.~M.~L. {Kappers}, ``The effect of global and local damping on the
  perception of hardness,'' \emph{IEEE Transactions on Haptics}, vol.~9, no.~3,
  pp. 409--420, 2016.

\bibitem{abidi2015haptics}
M.~H. Abidi, A.~Ahmad, S.~Darmoul, and A.~M. Al-Ahmari, ``Haptics assisted
  virtual assembly,'' \emph{IFAC-PapersOnLine}, vol.~48, no.~3, pp. 100--105,
  2015.

\bibitem{sagardia2016platform}
M.~Sagardia, T.~Hulin, K.~Hertkorn, P.~Kremer, and S.~Sch{\"a}tzle, ``A
  platform for bimanual virtual assembly training with haptic feedback in large
  multi-object environments,'' in \emph{Proc. ACM Conference on Virtual Reality
  Software and Technology}, 2016, pp. 153--162.

\bibitem{aggarwal2010training}
R.~Aggarwal, O.~T. Mytton, M.~Derbrew, D.~Hananel, M.~Heydenburg, B.~Issenberg,
  C.~MacAulay, M.~E. Mancini, T.~Morimoto, N.~Soper \emph{et~al.}, ``Training
  and simulation for patient safety,'' \emph{BMJ Quality \& Safety}, vol.~19,
  no. Suppl 2, pp. i34--i43, 2010.

\bibitem{zamani2019novel}
N.~Zamani, A.~Pourkand, C.~Salas, D.~M. Mercer, and D.~Grow, ``A novel approach
  for assessing and training the drilling skills of orthopaedic surgeons,''
  \emph{JBJS}, vol. 101, no.~16, p. e82, 2019.

\bibitem{WANG2019136}
D.~Wang, Y.~Guo, S.~Liu, Y.~Zhang, W.~Xu, and J.~Xiao, ``Haptic display for
  virtual reality: progress and challenges,'' \emph{Virtual Reality \&
  Intelligent Hardware}, vol.~1, no.~2, pp. 136--162, 2019.

\bibitem{Salisbury}
K.~Salisbury, D.~Brock, T.~Massie, N.~Swarup, and C.~Zilles,
  ``\BIBforeignlanguage{English (US)}{Haptic rendering: Programming touch
  interaction with virtual objects},'' in \emph{\BIBforeignlanguage{English
  (US)}{Proc. Symposium on Interactive 3D Graphics}}, Jan. 1995, pp. 123--130.

\bibitem{Diolaiti}
N.~Diolaiti, G.~Niemeyer, F.~Barbagli, and J.~Salisbury, ``A criterion for the
  passivity of haptic devices,'' in \emph{Proc. IEEE International Conference
  on Robotics and Automation}, vol. 2005, 05 2005, pp. 2452 -- 2457.

\bibitem{Kuchenbecker1}
K.~J. {Kuchenbecker}, J.~{Fiene}, and G.~{Niemeyer}, ``Improving contact
  realism through event-based haptic feedback,'' \emph{IEEE Transactions on
  Visualization and Computer Graphics}, vol.~12, no.~2, pp. 219--230, 2006.

\bibitem{Lawrence}
D.~A. {Lawrence}, L.~Y. {Pao}, A.~M. {Dougherty}, M.~A. {Salada}, and
  Y.~{Pavlou}, ``Rate-hardness: a new performance metric for haptic
  interfaces,'' \emph{IEEE Transactions on Robotics and Automation}, vol.~16,
  no.~4, pp. 357--371, 2000.

\bibitem{okamoto2012psychophysical}
S.~Okamoto, H.~Nagano, and Y.~Yamada, ``Psychophysical dimensions of tactile
  perception of textures,'' \emph{IEEE Transactions on Haptics}, vol.~6, no.~1,
  pp. 81--93, 2012.

\bibitem{wouter1}
W.~M.~B. Tiest, ``Tactual perception of material properties,'' \emph{Vision
  Research}, vol.~50, no.~24, pp. 2775--2782, 2010.

\bibitem{Roland}
R.~Harper and S.~S. Stevens, ``Subjective hardness of compliant materials,''
  \emph{Quarterly Journal of Experimental Psychology}, vol.~16, no.~3, pp.
  204--215, 1964.

\bibitem{Higashi1}
K.~{Higashi}, S.~{Okamoto}, Y.~{Yamada}, H.~{Nagano}, and M.~{Konyo},
  ``Hardness perception by tapping: Effect of dynamic stiffness of objects,''
  in \emph{Proc. IEEE World Haptics Conference}, 2017, pp. 37--41.

\bibitem{tiest2009cues}
W.~M.~B. Tiest and A.~M. Kappers, ``Cues for haptic perception of compliance,''
  \emph{IEEE Trans. on Haptics}, vol.~2, no.~4, pp. 189--199, 2009.

\bibitem{Friedman}
R.~Friedman, K.~Hester, B.~Green, and R.~Lamotte, ``Magnitude estimation of
  softness,'' \emph{Experimental Brain Research}, vol. 191, pp. 133--42, 09
  2008.

\bibitem{srinivasan1995tactual}
M.~A. Srinivasan and R.~H. LaMotte, ``Tactual discrimination of softness,''
  \emph{Journal of Neurophysiology}, vol.~73, no.~1, pp. 88--101, 1995.

\bibitem{Han}
G.~Han and S.~Choi, ``Extended rate-hardness: A measure for perceived
  hardness,'' in \emph{Proc. EuroHaptics Conference}, A.~M.~L. Kappers,
  J.~B.~F. van Erp, W.~M. Bergmann~Tiest, and F.~C.~T. van~der Helm, Eds.\hskip
  1em plus 0.5em minus 0.4em\relax Berlin, Heidelberg: Springer Berlin
  Heidelberg, 2010, pp. 117--124.

\bibitem{Okamura1}
A.~M. {Okamura}, M.~R. {Cutkosky}, and J.~T. {Dennerlein}, ``Reality-based
  models for vibration feedback in virtual environments,'' \emph{IEEE/ASME
  Transactions on Mechatronics}, vol.~6, no.~3, pp. 245--252, 2001.

\bibitem{Colgate1}
J.~E. {Colgate}, M.~C. {Stanley}, and J.~M. {Brown}, ``Issues in the haptic
  display of tool use,'' in \emph{Proc. IEEE/RSJ International Conference on
  Intelligent Robots and Systems, Human Robot Interaction and Cooperative
  Robots}, vol.~3, 1995, pp. 140--145 vol.3.

\bibitem{colgate1997passivity}
J.~E. Colgate and G.~G. Schenkel, ``Passivity of a class of sampled-data
  systems: Application to haptic interfaces,'' \emph{Journal of Robotic
  Systems}, vol.~14, no.~1, pp. 37--47, 1997.

\bibitem{hannaford2002time}
B.~Hannaford and J.-H. Ryu, ``Time-domain passivity control of haptic
  interfaces,'' \emph{IEEE Transactions on Robotics and Automation}, vol.~18,
  no.~1, pp. 1--10, 2002.

\bibitem{colgate1993implementation}
J.~E. Colgate, P.~E. Grafing, M.~C. Stanley, and G.~Schenkel, ``Implementation
  of stiff virtual walls in force-reflecting interfaces,'' in \emph{Proc. IEEE
  Virtual Reality Annual International Symposium}, 1993, pp. 202--208.

\bibitem{gillespie1996stable}
R.~B. Gillespie, M.~R. Cutkosky \emph{et~al.}, ``Stable user-specific haptic
  rendering of the virtual wall,'' in \emph{Proc. ASME International Mechanical
  Engineering Congress and Exhibition}, vol.~58, 1996, pp. 397--406.

\bibitem{Jong}
{Jong-Phil Kim} and {Jeha Ryu}, ``Energy bounding algorithm based on passivity
  theorem for stable haptic interaction control,'' in \emph{Proc. International
  Symposium on Haptic Interfaces for Virtual Environment and Teleoperator
  Systems}, 2004, pp. 351--357.

\bibitem{Singh}
H.~Singh, A.~Jafari, and J.-H. Ryu, ``Successive stiffness increment approach
  for high stiffness haptic interaction,'' in \emph{Haptics: Perception,
  Devices, Control, and Applications}, F.~Bello, H.~Kajimoto, and Y.~Visell,
  Eds.\hskip 1em plus 0.5em minus 0.4em\relax Cham: Springer International
  Publishing, 2016, pp. 261--270.

\bibitem{Singh2}
H.~{Singh}, A.~{Jafari}, and J.~{Ryu}, ``Increasing the rate-hardness of haptic
  interaction: Successive force augmentation approach,'' in \emph{Proc. IEEE
  World Haptics Conference}, 2017, pp. 653--658.

\bibitem{Han2}
G.~Han, S.~Jeon, and S.~Choi, ``Improving perceived hardness of haptic
  rendering via stiffness shifting: An initial study,'' in \emph{Proc. ACM
  Symposium on Virtual Reality Software and Technology}, New York, NY, USA,
  2009, p. 87–90.

\bibitem{Tiest1}
W.~M. Bergmann~Tiest and A.~M.~L. Kappers, ``Kinaesthetic and cutaneous
  contributions to the perception of compressibility,'' in \emph{Haptics:
  Perception, Devices and Scenarios}, M.~Ferre, Ed.\hskip 1em plus 0.5em minus
  0.4em\relax Berlin, Heidelberg: Springer Berlin Heidelberg, 2008, pp.
  255--264.

\bibitem{Park}
J.~{Park}, Y.~{Oh}, and H.~Z. {Tan}, ``Effect of cutaneous feedback on the
  perceived hardness of a virtual object,'' \emph{IEEE Transactions on
  Haptics}, vol.~11, no.~4, pp. 518--530, 2018.

\bibitem{Pourkand}
A.~{Pourkand} and J.~J. {Abbott}, ``Hybrid force-moment braking pulse: A haptic
  illusion to increase the perceived hardness of virtual surfaces,'' \emph{IEEE
  Robo. and Automation Letters}, vol.~5, no.~3, pp. 4588--4595, 2020.

\bibitem{Jeon}
S.~{Jeon}, ``Real stiffness augmentation for haptic augmented reality,''
  \emph{Presence}, vol.~20, no.~4, pp. 337--370, 2011.

\bibitem{Rodrigo}
V.~R. Mercado, M.~Marchal, and A.~Lecuyer, ``"haptics on-demand": A survey on
  encountered-type haptic displays,'' \emph{IEEE Transactions on Haptics}, pp.
  1--1, 2021.

\bibitem{snake}
B.~Araujo, R.~Jota, V.~Perumal, J.~X. Yao, K.~Singh, and D.~Wigdor, ``Snake
  charmer: Physically enabling virtual objects,'' in \emph{Proc. International
  Conference on Tangible, Embedded, and Embodied Interaction}, 2016, pp.
  218--226.

\bibitem{Mercado}
V.~Mercado, M.~Marchal, and A.~Lécuyer, ``Entropia: Towards infinite surface
  haptic displays in virtual reality using encountered-type rotating props,''
  \emph{IEEE Transactions on Visualization and Computer Graphics}, vol.~27,
  no.~3, pp. 2237--2243, 2021.

\bibitem{lamotte2000softness}
R.~H. LaMotte, ``Softness discrimination with a tool,'' \emph{Journal of
  Neurophysiology}, vol.~83, no.~4, pp. 1777--1786, 2000.

\bibitem{Ikeda}
Y.~Ikeda and S.~Hasegawa, ``Characteristics of perception of stiffness by
  varied tapping velocity and penetration in using event-based haptics,'' in
  \emph{Proc. Joint Virtual Reality Eurographics Conference on Virtual
  Environments}, 01 2009, pp. 113--116.

\bibitem{Higashi2}
K.~{Higashi}, S.~{Okamoto}, H.~{Nagano}, and Y.~{Yamada}, ``Effects of
  mechanical parameters on hardness experienced by damped natural vibration
  stimulation,'' in \emph{Proc. IEEE International Conference on Systems, Man,
  and Cybernetics}, 2015, pp. 1539--1544.

\bibitem{Culbertson17}
H.~Culbertson and K.~J. Kuchenbecker, ``Importance of matching physical
  friction, hardness, and texture in creating realistic haptic virtual
  surfaces,'' \emph{IEEE Trans. on Haptics}, vol.~10, no.~1, pp. 63--74, 2017.

\bibitem{zamani19}
N.~Zamani and H.~Culbertson, ``Effects of dental glove thickness on tactile
  perception through a tool,'' in \emph{2019 IEEE World Haptics Conference
  (WHC)}, 2019, pp. 187--192.

\bibitem{rank}
M.~Rank, T.~Schauss, A.~Peer, S.~Hirche, and R.~Klatzky, ``Masking effects for
  damping jnd,'' 06 2012, pp. 145--150.

\bibitem{zook2021effect}
Z.~Zook, J.~Fleck, and M.~K. O'Malley, ``Effect of tactile masking on
  multi-sensory haptic perception,'' \emph{IEEE Transactions on Haptics}, 2021.

\end{thebibliography}

\begin{IEEEbiography}[{\includegraphics[width=1in,height=1.25in,clip,keepaspectratio]{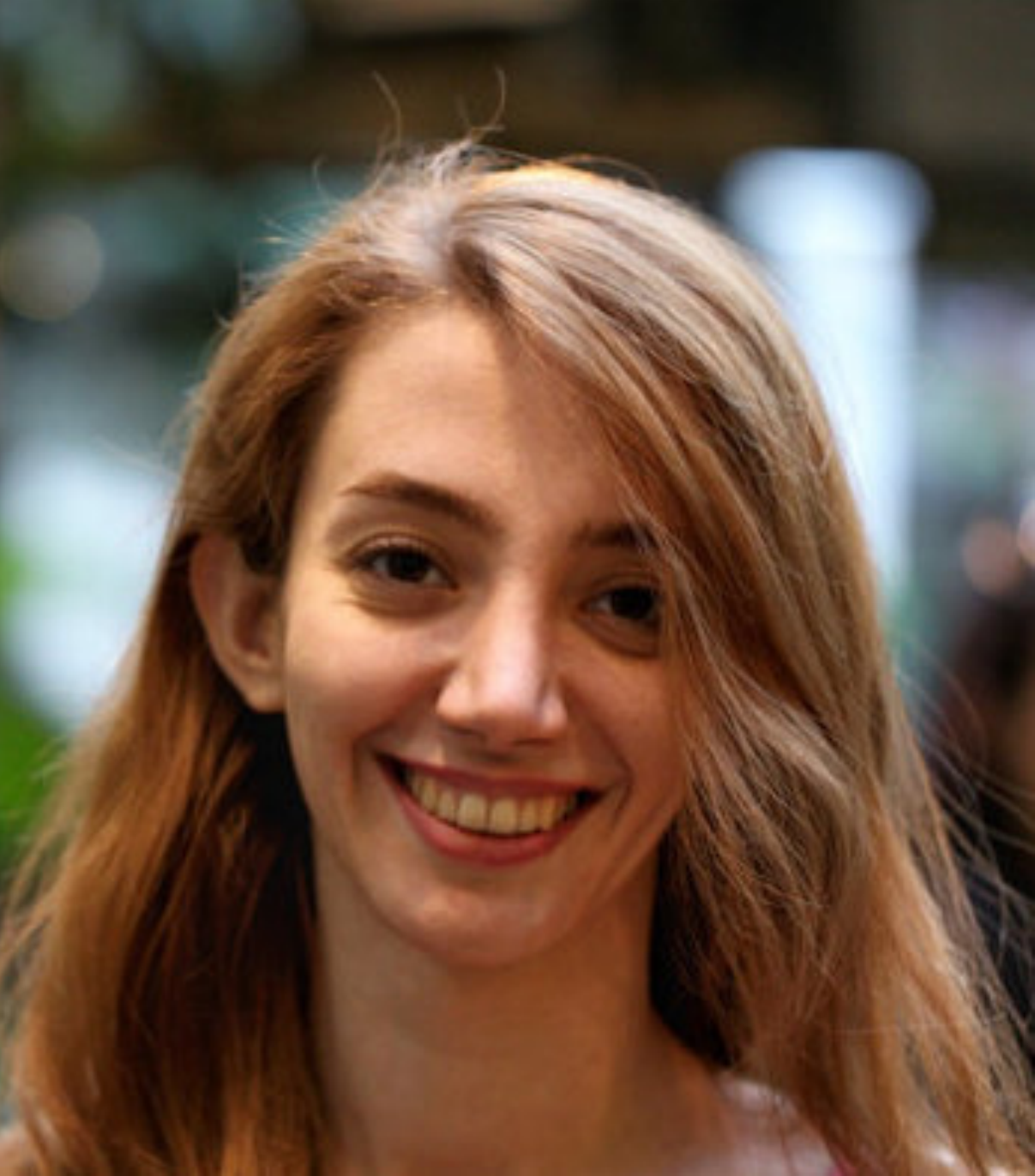}}]{Naghmeh Zamani}
Naghmeh Zamani is a Ph.D.\ student in the department of computer science at the University of Southern California since 2018. She received her M.S.\  degree in the department of mechanical engineering from New Mexico Tech, Socorro, New Mexico, in 2018. She earned the B.S.\ degree in Mechanical Engineering from the University of Kurdistan, Iran, in 2014.

\end{IEEEbiography}

\begin{IEEEbiography}[{\includegraphics[width=1in,height=1.25in,clip,keepaspectratio]{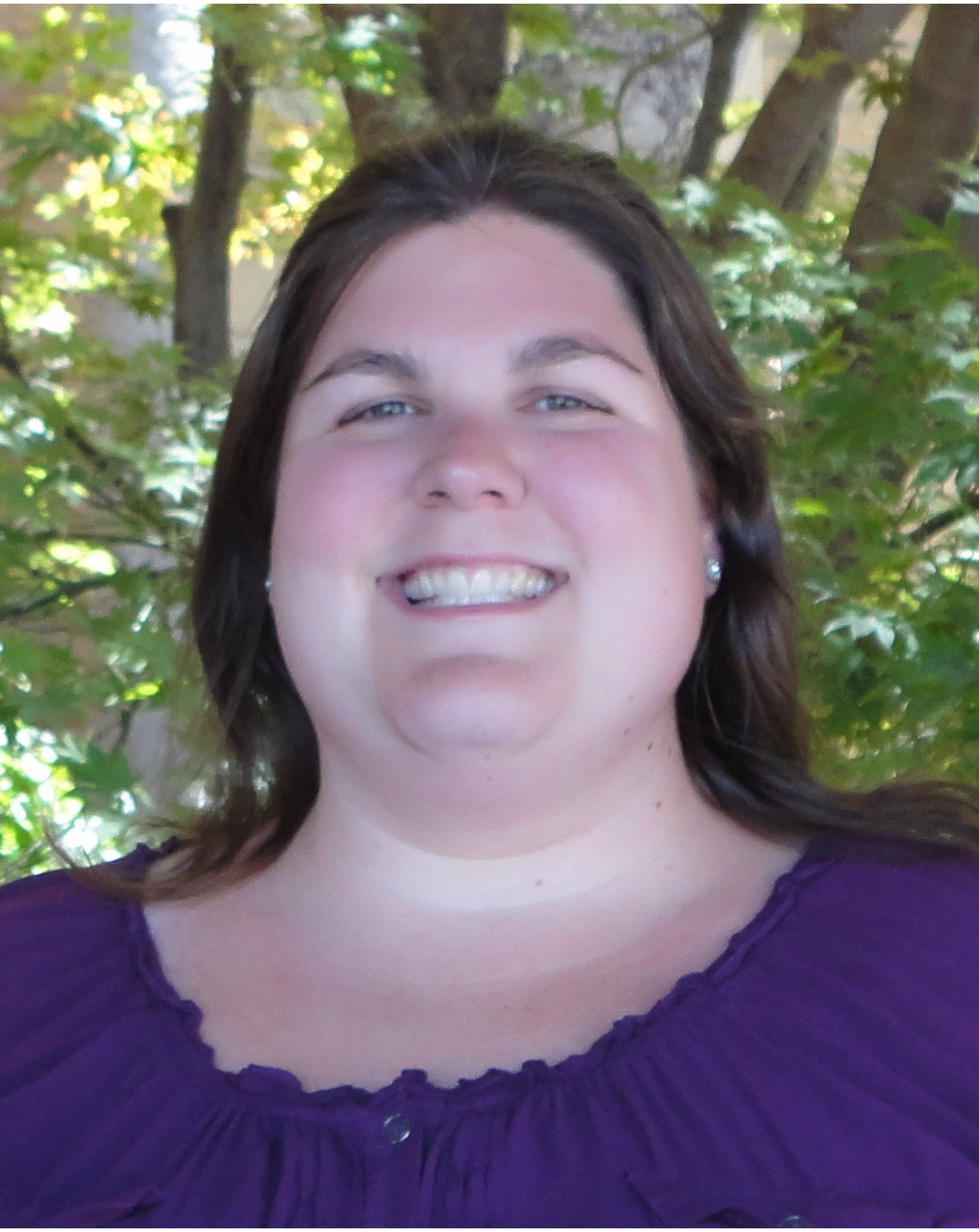}}]{Heather Culbertson}
Heather Culbertson is an assistant professor of Computer Science and Aerospace and Mechanical Engineering at the University of Southern California. She received the M.S.\ and Ph.D.\ degrees in the department of mechanical engineering and applied mechanics (MEAM) at the University of Pennsylvania in 2013 and 2015, respectively. She earned the B.S. degree in mechanical engineering at the University of Nevada, Reno in 2010. Prior to joining USC she was a research scientist at Stanford University. Her research focuses on the design and control of haptic devices and rendering systems, human-robot interaction, and virtual reality.
\end{IEEEbiography}

\end{document}